\pdfoutput=1
\documentclass[10pt,twocolumn,letterpaper]{article}

\usepackage{cvpr}
\usepackage{times}
\usepackage{epsfig}
\usepackage{graphicx}
\usepackage{amsmath}
\usepackage{amssymb}
\usepackage{multirow}
\usepackage{enumitem}
\usepackage{subcaption}
\usepackage{xcolor}
\usepackage{amsthm}
\usepackage{amsmath,amssymb}
\usepackage[ruled,vlined]{algorithm2e}
\usepackage{tabularx,booktabs}
\usepackage{hhline}
\newcolumntype{Y}{>{\centering\arraybackslash}X}

\usepackage[pagebackref=true,breaklinks=true,letterpaper=true,colorlinks,bookmarks=false]{hyperref}

\newtheorem{myprop}{Proposition}

\DeclareMathOperator*{\argmin}{arg\,min}
\DeclareMathOperator{\E}{\mathbb{E}}

\cvprfinalcopy 

\def\cvprPaperID{5075} 

\ifcvprfinal\fi
\begin{document}

\title{\ What Makes Training Multi-modal Classification Networks Hard?}

%

\author{%
  Weiyao Wang, Du Tran, Matt Feiszli \\
  Facebook AI \\
  \texttt{\{weiyaowang,trandu,mdf\}@fb.com} \\
}

\maketitle

\begin{abstract}

Consider end-to-end training of a multi-modal vs. a uni-modal network on a task with multiple input modalities: the multi-modal network receives more information, so it should match or outperform its uni-modal counterpart. In our experiments, however, we observe the opposite: the best uni-modal network often outperforms the multi-modal network. This observation is consistent across different combinations of modalities and on different tasks and benchmarks for video classification.

This paper identifies two main causes for this performance drop: first, multi-modal networks are often prone to overfitting due to their increased capacity. Second, different modalities overfit and generalize at different rates, so training them jointly with a single optimization strategy is sub-optimal. We address these two problems with a technique we call {Gradient-Blending}, which computes an optimal blending of modalities based on their overfitting behaviors. We demonstrate that Gradient Blending outperforms widely-used baselines for avoiding overfitting and achieves state-of-the-art accuracy on various tasks including human action recognition, ego-centric action recognition, and acoustic event detection.
\end{abstract}

\section{Introduction}
\label{sec:intro}





Consider a late-fusion multi-modal network, trained end-to-end to solve a task. Uni-modal solutions are a strict subset of the solutions available to the multi-modal network; a well-optimized multi-modal model should, in theory, always outperform the best uni-modal model. However, we show here that current techniques do not always achieve this. In fact, what we observe is contrary to common sense: the best uni-modal model often outperforms the joint model, across different modalities (Table~\ref{tab:of_naive}) and datasets (details in section~\ref{sec:ablation}). Anecdotally, the performance drop with multiple input streams appears to be common and was noted in~\cite{balanced_vqa_v2,AV_Dialogue,HypoOnlyHypoTesting,ShiftVQABaseline}. This (surprising) phenomenon warrants investigation and solution.

\begin{table}
    \small 
    \captionsetup{font=small}
    \begin{minipage}{1\linewidth}
        \centering
    	\begin{tabular}{|c|c c|c c|c|}
    	\hline
    	{\bf \footnotesize Dataset} & {\bf \footnotesize{Multi-modal} } & {\bf V@1} & {\bf \footnotesize{Best Uni}} & {\bf V@1} & {\bf Drop} \\
    	\hline
    	\multirow{4}{3em}{\footnotesize Kinetics} & A + RGB & 71.4 & RGB & \textbf{72.6} & -1.2\\
    	 & RGB + OF & 71.3 & RGB & \textbf{72.6} & -1.3 \\
    	 & A + OF & 58.3 & OF & \textbf{62.1} & -3.8 \\
    	 & \scriptsize{A + RGB + OF} & 70.0 & RGB & \textbf{72.6} & -2.6 \\
    	\hline
    	\end{tabular} 
    	\vspace{-3pt}
    	\caption{{\bf Uni-modal networks consistently outperform multi-modal networks.}  Best uni-modal networks vs late fusion multi-modal networks on Kinetics using video top-1 validation accuracy. Single stream modalities include video clips (RGB), Optical Flow (OF), and Audio (A). Multi-modal networks use the same architectures as uni-modal, with late fusion by concatenation at the last layer before prediction.} 
        \label{tab:of_naive}
    \end{minipage}
\vspace{-5mm} 
\end{table}



Upon inspection, the problem appears to be overfitting: multi-modal networks have higher train accuracy and lower validation accuracy. Late fusion audio-visual (A+RGB) network has nearly two times the parameters of a visual network, and one may suspect that the overfitting is caused by the increased number of parameters. 

\begin{figure}
\captionsetup{font=small}
\begin{center}
\includegraphics[width=1\linewidth]{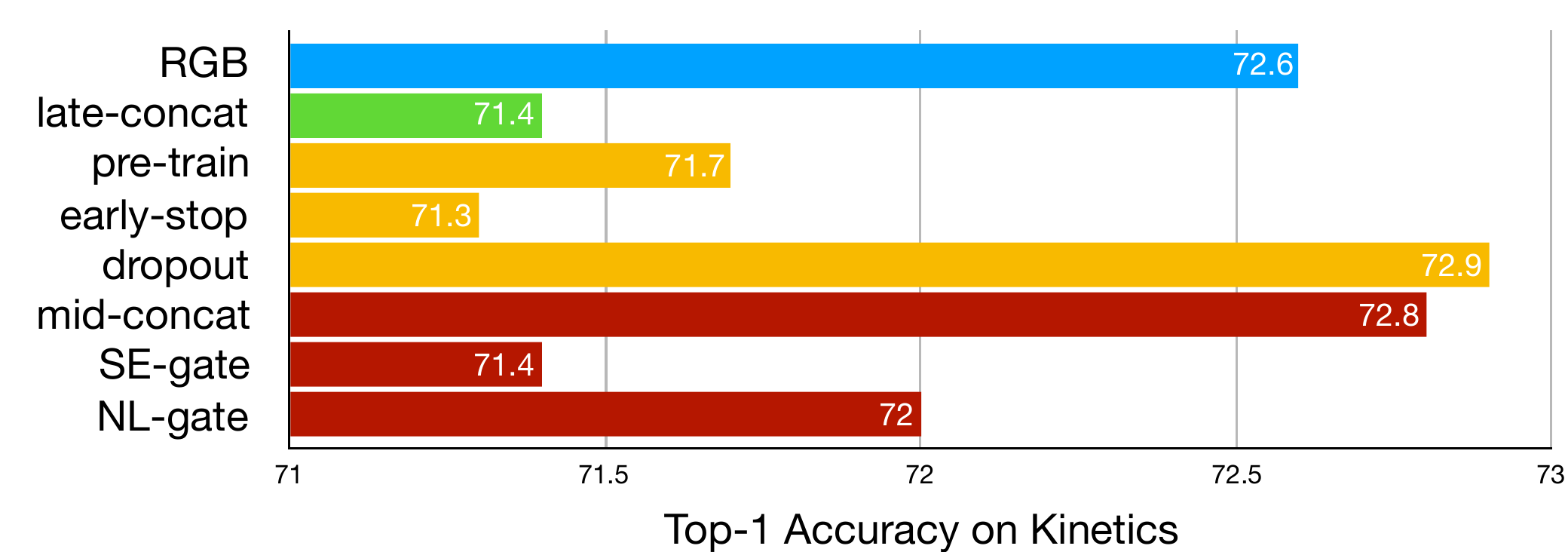}
\vspace{-6mm}
\caption{{\bf Standard regularizers do not provide a good improvement over the best Uni-modal network.} Best uni-modal network (RGB) vs standard approaches on a multi-modal network (RGB+Audio) on Kinetics.  Various methods to avoid overfitting (orange: Pre-training, Early-stopping, and Dropout) do not solve the issue. Different fusion architectures (red: Mid-concatenation fusion, SE-gate, and NL-gate) also do not help. Dropout and Mid-concatenation fusion approaches provide small improvements (+0.3\% and +0.2\%), while other methods degrade accuracy.}
\label{fig:diff_approach_fail}
\end{center}
\vspace{-10mm} 
\end{figure} 

There are two ways to approach this problem. First, one can consider solutions such as dropout~\cite{Dropout14}, pre-training, or early stopping to reduce overfitting. On the other hand, one may speculate that this is an architectural deficiency. We experiment with mid-level fusion by concatenation~\cite{Owens_2018_ECCV} and fusion by gating~\cite{Kiela18}, trying both Squeeze-and-Excitation (SE)~\cite{SENet} gates and Non-Local (NL)~\cite{XiaolongWang18} gates. 

Remarkably, none of these provide an effective solution. For each method, we record the best audio-visual results on Kinetics in Figure~\ref{fig:diff_approach_fail}. Pre-training fails to offer improvements, and early stopping tends to under-fit the RGB stream. Mid-concat and dropout provide only modest improvements over RGB model. We note that dropout and mid-concat (with 37\% fewer parameters compared to late-concat) make 1.5\% and 1.4\% improvements over late-concat, confirming the overfitting problem with late-concat. We refer to supplementary materials for details. 





How do we reconcile these experiments with previous multi-modal successes? Multi-modal networks have successfully been trained jointly on tasks including sound localization~\cite{ZhaoSOP18}, image-audio alignment~\cite{L3Net17}, and audio-visual synchronization~\cite{Owens_2018_ECCV,Korbar18}. However, these tasks cannot be performed with a single modality, so there is no uni-modal baseline and the performance drop found in this paper does not apply. In other work, joint training is avoided entirely by using pre-trained uni-modal features. Good examples include two-stream networks for video classification~\cite{SimonyanZ14,WangXW0LTG16,FeichtenhoferPZ16,I3D} and image+text classification~\cite{Arevalo17,Kiela18}. These methods do not train multiple modalities jointly, so they are again not comparable, and their accuracy may likely be sub-optimal due to independent training.






Our contributions in this paper include:
\begin{itemize}[nosep,leftmargin=0.6cm]
    \item We empirically demonstrate the significance of overfitting in joint training of multi-modal networks, and we identify two causes for the problem. We show the problem is architecture agnostic: different fusion techniques can also suffer the same overfitting problem. 
    \item We propose a metric to understand the problem quantitatively: the overfitting-to-generalization ratio (\emph{OGR}), with both theoretical and empirical justification.
    \item We propose a new training scheme which minimizes \emph{OGR} via an optimal blend (in a sense we make precise below) of multiple supervision signals. This {\bf Gradient-Blending} (G-Blend) method gives significant gains in ablations and achieves state-of-the-art (SoTA) accuracy on benchmarks including Kinetics, EPIC-Kitchen, and AudioSet by combining audio and visual signals. 
\end{itemize}

\noindent We note that G-Blend is task-agnostic, architecture-agnostic and applicable to other scenarios (e.g. used in~\cite{imvotenet} to combine point cloud with RGB for 3D object detection)

\subsection{Related Work} \label{sec:related_work}

\noindent \textbf{Video classification.} Video understanding has been one of the most active research areas in computer vision recently. There are two unique features with respect to videos: temporal information and multi-modality. Previous works have made significant progress in understanding temporal information~\cite{Karpathy14,Tran15,Wang_Transformation,P3D,Tran18,xie2017rethinking,slowfast}. However, videos are also rich in multiple modalities: RGB frames, motion vectors (optical flow), and audio. Previous works that exploit the multi-modal natures primarily focus on RGB+Optical Flow, with the creation of two-stream fusion  networks~\cite{SimonyanZ14,FeichtenhoferPZ16,FeichtenhoferNIPS16,WangXW0LTG16,I3D}, typically using pre-trained features and focusing on the fusion~\cite{Karpathy14,FeichtenhoferPZ16} or aggregation architectures~\cite{Ng15}. In contrast, we focus on joint training of the entire network. Instead of focusing on the architectural problem, we study model optimization: how to jointly learn and optimally blend multi-modal signals. With proper optimization, we show audio is useful for video classification.

\noindent \textbf{Multi-modal networks.} Our work is related to previous research on multi-modal networks~\cite{Baltruaitis2018MultimodalML} for classifications~\cite{SimonyanZ14,WangXW0LTG16,FeichtenhoferPZ16,Fukui16,I3D,Arevalo17,abs-1708-03805,Kiela18}, which primarily uses pre-training in contrast to our joint training. On the other hand, our work is related to cross-modal tasks~\cite{Weston:2011:WSU:2283696.2283856,NIPS2013_5204,Socher:2013:ZLT:2999611.2999716,VQA,balanced_binary_vqa,balanced_vqa_v2,ImageCaption16} and cross-modal self-supervised learning~\cite{ZhaoSOP18,L3Net17,Owens_2018_ECCV,Korbar18}. These tasks either take one modality as input and make prediction on the other modality (e.g. Visual-Q\&A~\cite{VQA,balanced_binary_vqa,balanced_vqa_v2}, image captioning~\cite{ImageCaption16}, sound localization~\cite{Owens_2018_ECCV,ZhaoSOP18} in videos) or uses cross-modality correspondences as self-supervision (e.g. image-audio correspondence~\cite{L3Net17}, video-audio synchronization~\cite{Korbar18}). Instead, we try to address the problem of joint training of multi-modal networks for classification. 

\noindent \textbf{Multi-task learning.} Our proposed Gradient-Blending training scheme is related to previous works in multi-task learning in using auxiliary loss~\cite{Kokkinos16,Eigen15,Kendall18,GradNorm18}. These methods either use uniform/manually tuned weights, or learn the weights as parameters during training (no notion of overfitting prior used), while our work re-calibrates supervision signals using \textit{a prior} \emph{OGR}.

\section{Multi-modal training via Gradient-Blending} 
\label{sec:technical_approach}

\subsection{Background} \label{sec:technical_approach}

\begin{figure*}[h!]
\captionsetup{font=footnotesize}
\begin{center}
\includegraphics[width=0.65\linewidth]{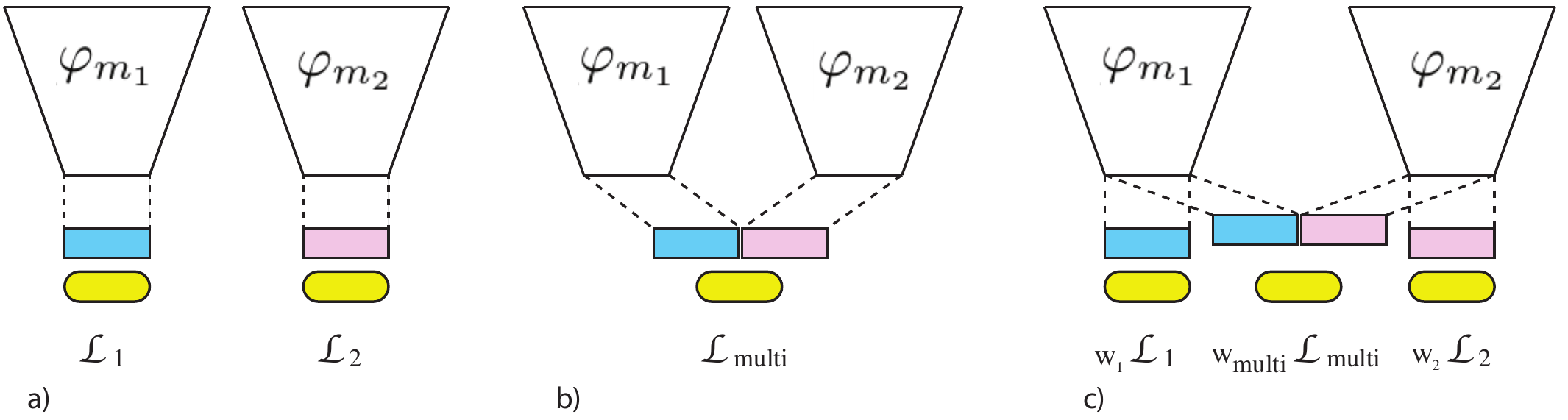}
\vspace{-3mm} 
\caption{{\bf Uni- vs. multi-modal joint training}. a) Uni-modal training of two different modalities. b) Naive joint training of two modalities by late fusion. c) Joint training of two modalities with weighted blending of supervision signals. Different deep network encoders (white trapezoids) produce features (blue or pink rectangles) which are concatenated and passed to a classifier (yellow rounded rectangles).}
\label{fig:av_joint_training}
\end{center}
\vspace{-8mm} 
\end{figure*} 

\noindent {\bf Uni-modal network}. Given train set $\mathcal{T} = \{X_{1 \dots n}, y_{1 \dots n}\}$, where $X_i$ is the $i$-th training example and $y_i$ is its true label, training on a single modality $m$ (e.g. RGB frames, audio, or optical flows) means minimizing an empirical loss:
\begin{equation}
    \mathcal{L}\left(\mathcal{C}\left(\varphi_m(X)\right), y\right)
\label{equ:single}
\end{equation}
where $\varphi_m$ is normally a deep network with parameter $\Theta_m$, and $\mathcal{C}$ is a classifier, typically one or more fully-connected (FC) layers with parameter $\Theta_c$. For classification problems considered here, $\mathcal{L}$ is the cross entropy loss. Minimizing Eq.~\ref{equ:single} gives a solution $\Theta_m^*$ and $\Theta_c^*$. Fig.~\ref{fig:av_joint_training}a shows independent training of two modalities $m_1$ and $m_2$.

\noindent {\bf Multi-modal network}. We train a late-fusion model on $M$ different modalities ($\{m_i\}_1^k$). Each modality is processed by a different deep network $\varphi_{m_i}$ with parameter $\Theta_{m_i}$, and their features are fused and passed to a classifier $\mathcal{C}$. Formally, training is done by minimizing the loss:
\begin{equation}
    \resizebox{0.9\hsize}{!}{$\mathcal{L}_{multi} = \mathcal{L}\left(\mathcal{C}\left(\varphi_{m_1} \oplus \varphi_{m_2} \oplus \dots \oplus \varphi_{m_k}\right), y\right)$} 
\label{equ:multi}
\end{equation}
where $\oplus$ denotes a fusion operation (e.g. concatenation). Fig.~\ref{fig:av_joint_training}b shows an example of a joint training of two modalities $m_1$ and $m_2$. The multi-modal network in Eq.~\ref{equ:multi} is a super-set of the uni-model network in Eq.~\ref{equ:single}: for any solution to Eq.~\ref{equ:single} on any modality $m_i$, one can construct an equally-good solution to Eq.~\ref{equ:multi} by choosing parameters $\Theta_c$ that mute all modalities other than $m_i$. In practice, this solution is not found, and we next explain why.

\subsection{Generalizing \emph{vs.} Overfitting}
\label{subsec:learning_vs_overfitting}

\begin{figure}
\captionsetup{font=small}
\begin{center}
\includegraphics[width=0.7\linewidth]{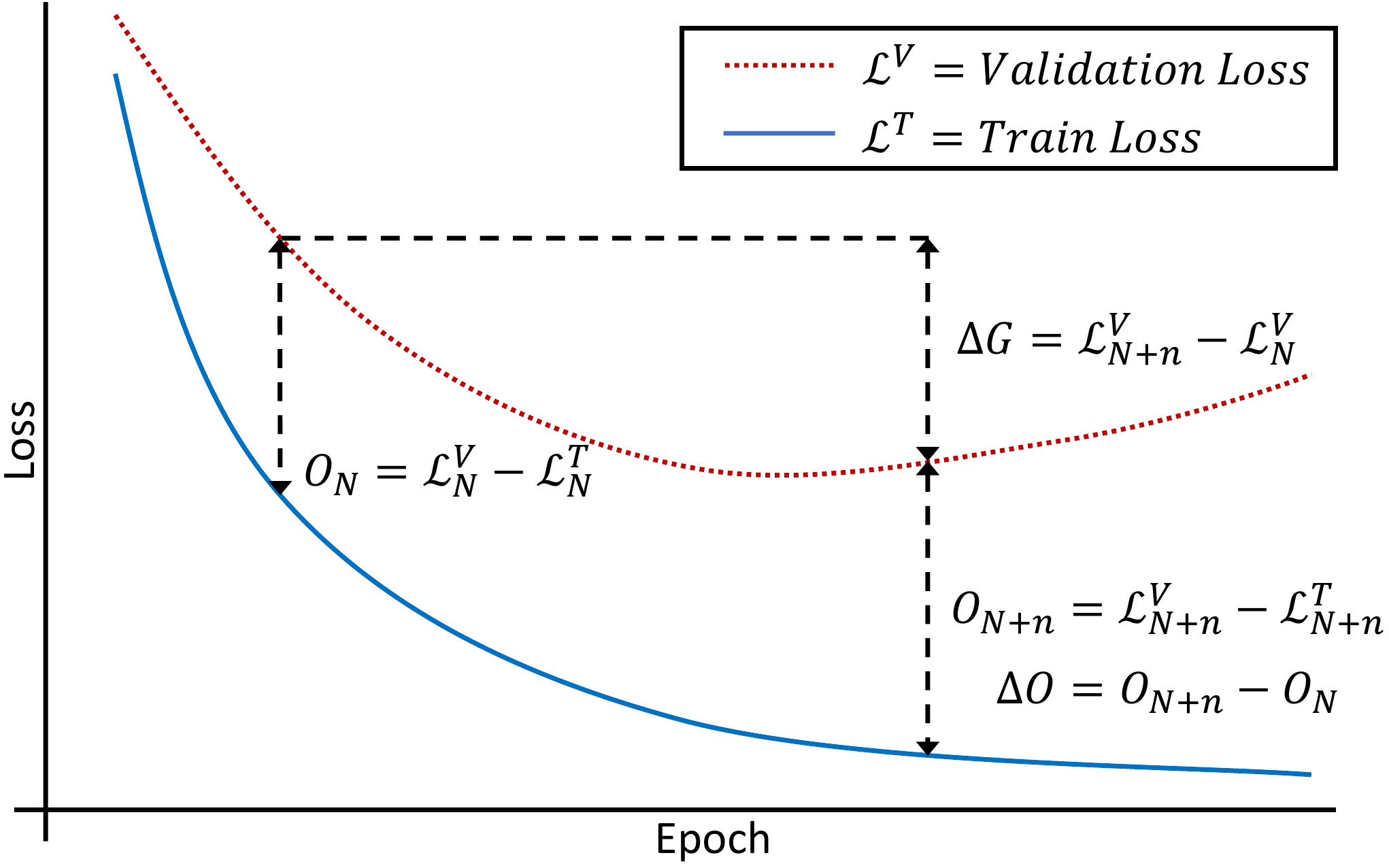}
\caption{{\bf Overfitting-to-Generalization Ratio.} Between any two training checkpoints, we can measure the change in overfitting and generalization. When $\frac{\Delta O}{\Delta V}$ is small, the network is learning well and not overfitting much.}
\label{fig:OGR_demo}
\end{center}
\vspace{-8mm} 
\end{figure}

Overfitting is typically understood as learning patterns in a train set that do not generalize to the target distribution. Given model parameters at epoch $N$, let $\mathcal{L}^{\mathcal{T}}_N$ be the model's average loss over the fixed train set, and $\mathcal{L}^*_N$ be the ``true'' loss w.r.t the hypothetical target distribution. (In what follows, $\mathcal{L}^*$ is approximated by a held-out validation loss $\mathcal{L}^{\mathcal{V}}$.)  We define overfitting at epoch $N$ as the gap between $\mathcal{L}^{\mathcal{T}}_N$ and $\mathcal{L}^*_N$ (approximated by $O_N$ in fig.~\ref{fig:OGR_demo}). The quality of training between two model checkpoints can be measured by the change in overfitting and generalization ($\Delta O$, $\Delta G$ in fig.~\ref{fig:OGR_demo}). Between checkpoints $N$ and $N+n$, we can define the overfitting-to-generalization-ratio ($OGR$): 

\begin{equation}
   OGR \equiv \left| \frac{\Delta O_{N,n}}{\Delta G_{N,n}} \right| = \left|  \frac{O_{N+n}-O_N}{\mathcal{L}^*_N - \mathcal{L}^*_{N+n}} \right|
\label{equ:OGR_def}
\end{equation}

$OGR$ between checkpoints measures the quality of learned information (with cross-entropy loss, it is the ratio of bits not generalizable to bits which do generalize). We propose minimizing $OGR$ during training. However, optimizing $OGR$ globally would be very expensive (e.g. variational methods over the whole optimization trajectory). In addition, very underfit models, for example, may still score quite well (difference of train loss and validation loss is very small for underfitting models; in other words, $O$ is small).  

Therefore, we propose to solve an infinitesimal problem: given several estimates of the gradient, blend them to minimize an infinitesimal $OGR^2$. We  apply this blend to our optimization process (e.g. SGD with momentum). Each gradient step now increases generalization error as little as possible per unit gain on the validation loss, minimizing overfitting. In a multi-modal setting, this means we combine gradient estimates from multiple modalities and minimize $OGR^2$ to ensure each gradient step now produces a gain no worse than that of the single best modality. As we will see in this paper, this $L^2$ problem admits a simple, closed-form solution, is easy to implement, and works well in practice.







Consider a single parameter update step with estimate $\hat{g}$ for the gradient. As the distance between two checkpoints is small (in the neighborhood in which a gradient step is guaranteed to decrease the train loss), we use the first-order approximations: $\Delta G\approx \langle \nabla \mathcal{L}^*, \hat{g} \rangle$ and $\Delta O\approx \langle \nabla \mathcal{L}^{\mathcal{T}}- \mathcal{L}^*, \hat{g} \rangle$. Thus, $OGR^2$ for a single vector $\hat{g}$ is
\begin{equation}
    OGR^2 = \left( \frac{ \langle \nabla \mathcal{L}^{\mathcal{T}} - \nabla \mathcal{L}^*, \hat{g} \rangle}{\langle \nabla \mathcal{L}^*, \hat{g} \rangle} \right)^2 
\label{equ:OLR}
\end{equation}

See supplementary materials for details on $OGR$.

\subsection{Blending of Multiple Supervision Signals by \textit{OGR} Minimization} \label{subsec:opt_blend_OLR}


We can obtain multiple estimates of gradient by attaching classifiers to each modality's features and to the fused features (see fig \ref{fig:av_joint_training}c). Per-modality gradient $\{\hat{g}_i\}_{i=1}^k$ are obtained by back-propagating through each loss separately (so per-modality gradients contain many zeros in other parts of the network). Our next result allows us to blend them all into a single vector with better generalization behavior.

\begin{myprop}[Optimal Gradient Blend]
Let $\{v_k\}_{0}^M$ be a set of estimates for $\nabla \mathcal{L}^*$ whose overfitting satisfies $\E \left[\langle \nabla \mathcal{L}^{\mathcal{T}} - \nabla \mathcal{L}^*, v_k \rangle \langle \nabla \mathcal{L}^{\mathcal{T}} - \nabla \mathcal{L}^*, v_j \rangle \right] = 0$ for $j \ne k$.  Given the constraint $\sum_k w_k = 1$ the optimal weights $w_k \in \mathbb{R}$ for the problem
\begin{equation}
    w^* = \argmin_{w}\E\left[\left(\frac{\langle \nabla \mathcal{L}^{\mathcal{T}} - \nabla \mathcal{L}^*, \sum_k w_k v_k \rangle}{\langle \nabla \mathcal{L}^*, \sum_k w_k v_k \rangle}\right)^2\right]
\label{equ:weight_argmin}
\end{equation}
are given by 
\begin{equation}
    w_k^* = \frac{1}{Z} \frac{\langle \nabla \mathcal{L}^*, v_k \rangle}{\sigma_k^2}
\label{equ:optimal_weight}
\end{equation}
where $\sigma_k^2 \equiv \E[\langle \nabla \mathcal{L}^{\mathcal{T}} - \nabla \mathcal{L}^*, v_k \rangle^2]$ and $Z = \sum_k \frac{\langle \nabla \mathcal{L}^*, v_k \rangle}{2\sigma_k^2}$ is a normalizing constant.
\label{prop:optimal_blend}
\end{myprop}

Assumption $\E \left[\langle \nabla \mathcal{L}^{\mathcal{T}} - \nabla \mathcal{L}^*, v_k \rangle \langle \nabla \mathcal{L}^{\mathcal{T}} - \nabla \mathcal{L}^*, v_j \rangle \right]=0$ will be false when two models' overfitting is very correlated. However, if this is the case then very little can be gained by blending their gradients. In informal experiments we have indeed observed that these cross terms are often small relative to the $\E \left[\langle \nabla \mathcal{L}^{\mathcal{T}} - \nabla \mathcal{L}^*, v_k \rangle^2 \right]$.  This is likely due to complementary information across modalities, and we speculate that this happens naturally as joint training tries to learn complementary features across neurons. Please see supplementary materials for proof of Proposition~\ref{prop:optimal_blend}, including formulas for the correlated case.

Proposition~\ref{prop:optimal_blend} may be compared with well-known results for blending multiple estimators; e.g. for the mean, a minimum-variance estimator is obtained by blending uncorrelated estimators with weights inversely proportional to the individual variances (see e.g. \cite{StatsBlend}). Proposition 1 is similar, where variance is replaced by $O^2$ and weights are inversely proportional to the individual $O^2$ (now with a numerator $G$).

\subsection{Use of \emph{OGR} and Gradient-Blending in practice}
\label{subsec:OLR_in_practice}
We adapt a multi-task architecture to construct an approximate solution to the optimization above (fig \ref{fig:av_joint_training}c). 


{\bf Optimal blending by loss re-weighting} 
At each back-propagation step, the per-modality gradient for $m_i$ is $\nabla \mathcal{L}_{i}$, and the gradient from the fused loss is given by Eq.~\ref{equ:multi} (denote as $\nabla \mathcal{L}_{k+1}$). Taking the gradient of the blended loss
\begin{equation}
    \mathcal{L}_{blend} = \sum_{i=1}^{k+1} w_{i} \mathcal{L}_{i}
\label{equ:blend_loss}
\end{equation}
thus produces the blended gradient $\sum_{i=1}^{k+1} w_{i}\nabla \mathcal{L}_{i}$. For appropriate choices of $w_i$ this yields a convenient way to implement gradient blending. Intuitively, loss reweighting re-calibrates the learning schedule to balance the generalization/overfitting rate of different modalities. 

{\bf Measuring \emph{OGR} in practice.} In practice, $\nabla \mathcal{L}^*$ is not available. To measure \emph{OGR}, we hold out a subset $\mathcal{V}$ of the training set to approximate the true distribution (i.e. $\mathcal{L}^\mathcal{V} \approx \mathcal{L}^*$). We find it is equally effective to replace the loss measure by an accuracy metric to compute $G$ and $O$ and estimate optimal weights from Gradient-Blending. To reduce computation costs, we note that weights estimation can be done on a small subset of data, without perturbing the weights too much (see supplementary materials).

Gradient-Blending algorithms take inputs of training data $\mathcal{T}$, validation set $\mathcal{V}$, $k$ input modalities $\{m_i\}_{i=1}^k$ and a joint head $m_{k+1}$ (Fig.~\ref{fig:av_joint_training}c). In practice we can use a subset of training set $\mathcal{T}'$ to measure train loss/ accuracy. To compute the Gradient-Blending weights when training from $N$ for $n$ epochs, we provide a Gradient-Blending weight estimation in Algorithm~\ref{algo:g_b_subroutine}. We propose two versions of gradient-blending:
\begin{enumerate}[nosep]
    \item \textbf{Offline Gradient-Blending} is a simple version of gradient-blending. We compute weights only once, and use a fixed set of weights to train entire epoch. This is very easy to implement. See Algorithm~\ref{algo:g_b_offline}.
    \item \textbf{Online Gradient-Blending} is the full version. We re-compute weights regularly (e.g. every $n$ epochs -- called a \textit{super-epoch}), and train the model with new weights for a super-epoch. See Algorithm~\ref{algo:g_b_online}.
\end{enumerate}
Empirically, offline performs remarkably well. We compare the two in section~\ref{sec:ablation}, with online giving additional gains.


\vspace*{-3mm}
\begin{algorithm}
\SetAlgoLined
\SetKwInOut{Input}{input}
\Input{$\varphi^N$, \quad Model checkpoint at epoch $N$ \newline 
$n$, \quad \# of epochs}
\KwResult{A set of optimal weights with for $k+1$ losses.}
 \For{$i=1,...,k+1$}{
    Initialize uni-modal/ naive multi-modal network $\varphi^N_{m_i}$ from corresponding parameters in $\varphi^N$ \;
    Train $\varphi^N_{m_i}$ for $n$ epochs on $\mathcal{T}$, resulting model $\varphi^{N+n}_{m_i}$\;
    Compute amount of overfitting $O^i=O_{N,n}$, generalization $G^i=G_{N,n}$ according to Eq.\ref{equ:OGR_def} using $\mathcal{V}$ and $\mathcal{T'}$ for modality $m_i$\;
 }
 Compute a set of loss $\{w_i^*\}_{i=1}^{k+1}=\frac{1}{Z}\frac{G^i}{{O^i}^2}$  \;
 \caption{G-B Weight Estimation: $GB\_Estimate$}
\label{algo:g_b_subroutine}
\end{algorithm}

\vspace*{-6mm}

\begin{algorithm}
\SetAlgoLined
\SetKwInOut{Input}{input}
\Input{$\varphi^0$, \quad Initialized model \newline 
$N$, \quad \# of epochs}
\KwResult{Trained multi-head model $\varphi^N$}
Compute per-modality weights $\{w_i\}_{i=1}^k=GB\_Estimate(\varphi^0,N)$ \;
Train $\varphi^0$ with $\{w_i\}_{i=1}^k$ for $N$ epochs to get $\varphi^N$ \;
\caption{Offline Gradient-Blending}
\label{algo:g_b_offline}
\end{algorithm}

\vspace*{-6mm}

\begin{algorithm}
\SetAlgoLined
\SetKwInOut{Input}{input}
\Input{$\varphi^0$, \quad Initialized model \newline 
$N$, \quad \# of epochs \newline 
$n$, \quad super-epoch length}
\For{$i=0,...,\frac{N}{n}-1$}{
    Current epoch $N_i=i*n$ \;
    Compute per-modality weights $\{w_i\}_{i=1}^k=GB\_Estimate(\varphi^{N_i},N_i+n)$ \;
    Train $\varphi^{N_i}$ with $\{w_i\}_{i=1}^k$ for $n$ epochs to $\varphi^{N_i+n}$ \;
}
\caption{Online Gradient-Blending}
\label{algo:g_b_online}
\end{algorithm}

\vspace*{-3mm}

\section{Ablation Experiments}
\label{sec:ablation}

\subsection{Experimental setup} \label{subsec:ablation_setup}

\noindent {\bf Datasets.} We use three video datasets for ablations: Kinetics, mini-Sports, and mini-AudioSet. {\bf Kinetics} is a standard benchmark for action recognition with 260k videos~\cite{Kinetics} of $400$ human action classes. We use the train split (240k) for training and the validation split (20k) for testing. {\bf Mini-Sports} is a subset of Sports-1M~\cite{Karpathy14}, a large-scale classification dataset with 1.1M videos of 487 different fine-grained sports. We uniformly sampled 240k videos from train split and 20k videos from the test split. {\bf Mini-AudioSet} is a subset of AudioSet~\cite{audioset}, a multi-label dataset consisting of 2M videos labeled by 527 acoustic events. AudioSet is very class-unbalanced, so we remove tiny classes and subsample the rest (see supplementary). The balanced mini-AudioSet has 418 classes with 243k videos.


\noindent {\bf Input preprocessing \& augmentation.} We consider three modalities: RGB, optical flow and audio. For RGB and flow, we use input clips of 16$\times$224$\times$224 as input. We follow~\cite{CSN19} for visual pre-processing and augmentation. For audio, we use log-Mel with $100$ temporal frames by $40$ Mel filters. Audio and visual are temporally aligned.

\noindent \textbf{Backbone architecture.} We use \emph{ResNet3D}~\cite{Tran18} as our visual backbone for RGB and flow and \emph{ResNet}~\cite{KaimingHe16} as our audio model, both with 50 layers. For fusion, we use a two-FC-layer network on concatenated features from visual and audio backbones, followed by one prediction layer. 



\noindent {\bf Training and testing.} We train our models with synchronous distributed SGD on GPU clusters using Caffe2~\cite{caffe2}, with setup as~\cite{Tran18}. We hold out a small portion of training data for weight estimate (8\% for Kinetics and mini-Sports, 13\% for mini-AudioSet). The final video prediction is made by using center crops of 10 uniformly-sampled clips and averaging the 10 predictions.

\subsection{Overfitting Problems in Naive Joint Training} \label{subsec:Exp_of}

\begin{figure*}[h!]
    \captionsetup{font=footnotesize}
\begin{center}
\includegraphics[width=0.65\linewidth]{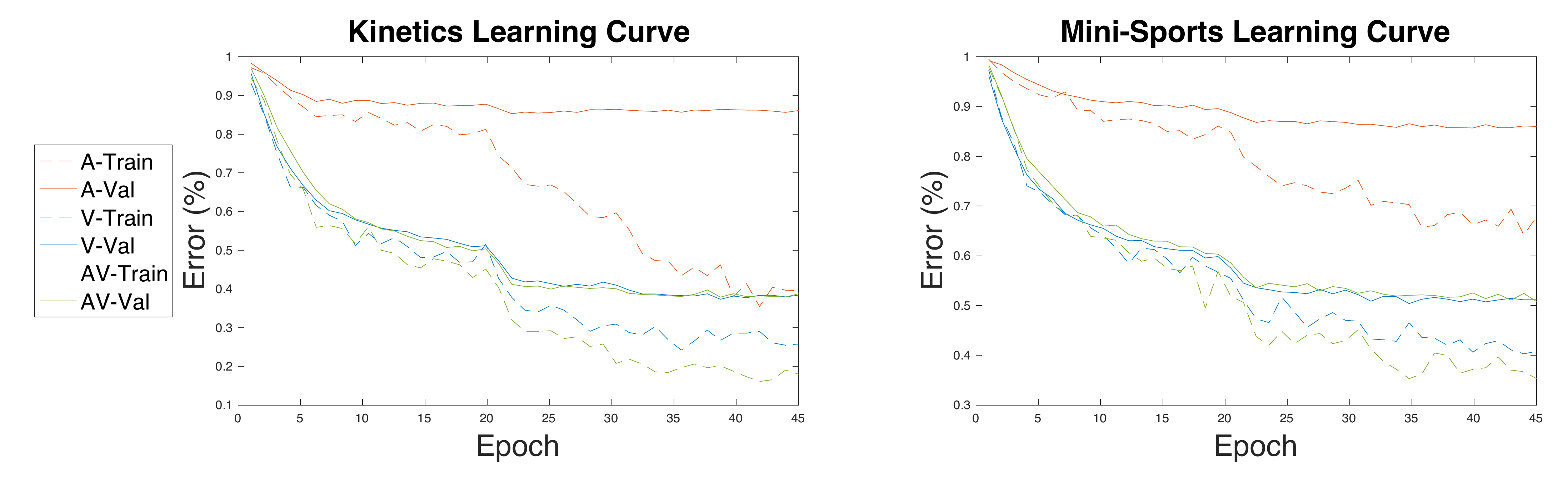}
\vspace{-3mm} 
\caption{\textbf{Severe overfitting of naive audio-video models on Kinetics and mini-Sports.} The learning curves (error-rate) of audio model (A), video model (V), and the naive joint audio-video (AV) model on Kinetics (left) and mini-Sports (right). Solid lines plot validation error while dashed lines show train error. The audio-video model overfits more than visual model and is inferior to the video-only model on validation loss.}
\label{fig:lr_of}
\end{center}
\vspace{-8mm} 
\end{figure*}

We first compare naive audio-RGB joint training with unimodal audio-only and RGB-only training. Fig.~\ref{fig:lr_of} plots the training curves on Kinetics (left) and mini-Sports (right). On both datasets, the audio model overfits the most and video overfits least.  The naive joint audio-RGB model has lower training error and higher validation error compared with the video-only model; i.e. naive audio-RGB joint training increases overfitting, explaining the accuracy drop compared to video alone.



We extend the analysis and confirm severe overfitting on other multi-modal problems. We consider all 4 possible combinations of the three modalities (audio, RGB, and optical flow). In every case, the validation accuracy of naive joint training is significantly worse than the best single stream model (Table~\ref{tab:of_naive}), and training accuracy is almost always higher (see supplementary materials).

\subsection{Gradient-Blending is an effective regularizer}


In this ablation, we first compare the performance of online and offline versions of G-Blend. Then we show that G-Blend works with different types of optimizers, including ones with adaptive learning rates. Next, we show G-Blend improves the performance on different multi-modal problems (different combinations of modalities), different model architectures and different tasks. 

\begin{figure}
 \captionsetup{font=footnotesize}
 \begin{subfigure}[b]{0.49\linewidth}
    \captionsetup{font=footnotesize}
    {\includegraphics[width=\linewidth]{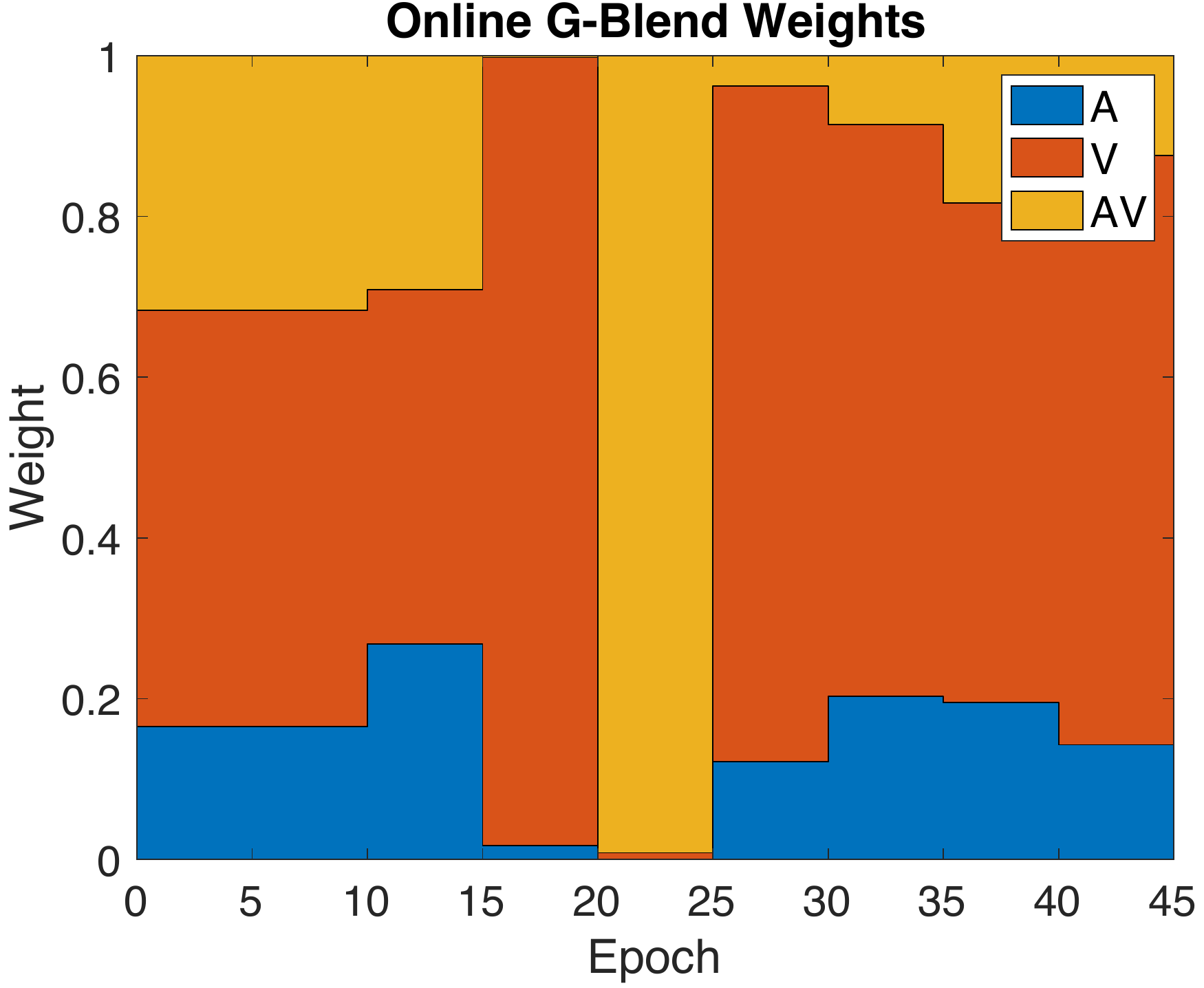}}
    \vspace{-5mm} 
    \caption{}
 \end{subfigure}
 \begin{subfigure}[b]{0.49\linewidth}
    \captionsetup{font=footnotesize}
    {\includegraphics[width=\linewidth]{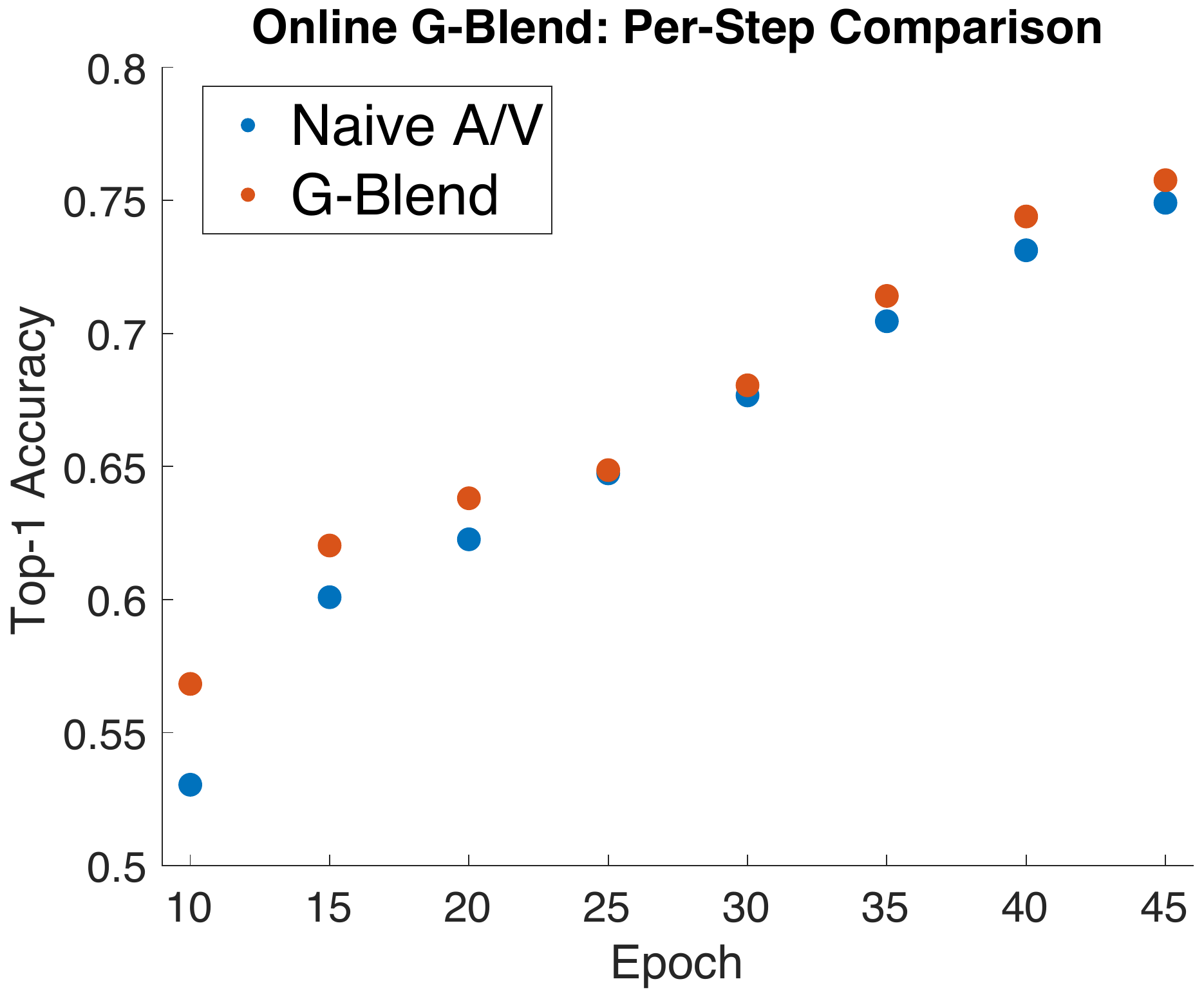}}
    \vspace{-5mm} 
    \caption{}
 \end{subfigure}
 \vspace{-3mm} 
 \caption{\textbf{Online G-Blend.} {\bf (a) Online G-Blend weights for each head.} {\bf (b) Online G-Blend outperforms naive training on each super-epoch}. For each super-epoch (5 epochs), we use the same snapshot of the model learned by G-Blend, and compare the performance of the models trained by G-Blend and naive at the next 5 epochs. G-Blend always outperforms naive training. This proves that G-Blend always learn more generalize information at a per-step level.}
 \vspace{-5mm} 
 \label{fig:online_g_b}
\end{figure}

\noindent \textbf{Online G-Blend Works.} We begin with the complete version of G-Blend, online G-Blend. We use an initial super-epoch size of 10 (for warmup), and a super-epoch size of 5 thereafter. On Kinetics with RGB-audio setting, online Gradient-Blending surpasses both uni-modal and naive multi-modal baselines, by 3.2\% and 4.1\% respectively. The weights for online are in fig.~\ref{fig:online_g_b}a. In general, weights tend to be stable at first with slightly more focused on visual; then we see a transition at epoch 15 where the model does ``pre-training'' on visual trunk; at epoch 20 A/V trunk got all weights to sync the learning from visual trunk. After that, weights gradually stabilize again with a strong focus on visual learning. We believe that, in general, patterns learned by neural network are different at different stage of training (e.g.\cite{Nakkiran2019SGDON}), thus the overfitting / generalization behavior also changes during training; this leads to different weights at different stages of the training.

Moreover, we observe that G-Blend always outperforms naive training in the online setting (Fig.~\ref{fig:online_g_b}b). With the same initialization (model snapshots at epoch 0,10,15,...,40), we compare the performance of G-Blend model and naive training after a super-epoch (at epoch 10,15,20,...,45), and G-Blend models always outperform naive training. This shows that G-Blend always provides more generalizable training information, empirically proving proposition~\ref{prop:optimal_blend}. Additionally, it shows the relevance of minimizing $OGR$, as using weights that minimize $OGR$ improves performance of the model. For fair comparison, we fix the main trunk and finetune the classifier for both Naive A/V and G-Blend as we want to evaluate the quality of their backbones. At epoch 25, the gain is small since G-Blend puts almost all weights on A/V head, making it virtually indistinguishable from naive training for that super-epoch.

\begin{table}[]
    \centering
    \captionsetup{font=footnotesize}
    {\small
    	\begin{tabular}{|c|c|c|c|}
        	\hline
        	Method & Clip & V@1 & V@5  \\
        	\hline
        	Naive Training & 61.8 & 71.7 & 89.6 \\
        	RGB Only & 63.5 & 72.6 & 90.1 \\
        	\hline
        	Offline G-Blend & 65.9 & 74.7 & 91.5 \\
        	\hline
        	Online G-Blend & \bf 66.9 & \bf 75.8 & \bf 91.9 \\
        	\hline
    	\end{tabular}
	}
	\vspace{-3pt}
	\caption{\textbf{Both offline and online Gradient-Blending outperform Naive late fusion and RGB only.} Offline G-Blend is lightly less accurate compared with the online version, but much simpler to implement.} 
	\label{tab:online}
	\vspace{-3mm} 
\end{table}

\noindent \textbf{Offline G-Blend Also Works.} Although online G-Blend gives significant gains and addresses overfitting well, it is more complicated to implement, and somewhat slower due to the extra weight computations. As we will now see, Offline G-Blend can be easily adopted and works remarkably well in practice. On the same audio-RGB setting on Kinetics, offline G-Blend also outperforms uni-modal baseline and naive joint training by a large margin, 2.1\% and 3.0\% respectively (Table~\ref{tab:online}), and is only slightly worse than online (-1.1\%). Based on such observation, we opt to use offline G-Blend in the rest of the ablations, demonstrating its performance across different scenarios. We speculate the online version will be particularly useful for some cases not covered here, for example a fast-learning low-capacity model (perhaps using some frozen pre-trained features), paired with a high-capacity model trained from scratch.

\noindent \textbf{Adaptive Optimizers.} Section~\ref{subsec:learning_vs_overfitting} introduced G-Blend in an infinitesimal setting: blending different gradient estimation at a single optimization step and assumes same learning rate for each gradient estimator. This is true for many popular SGD-based algorithms, such as SGD with Momentum. However, the assumption may not be rigorous with adaptive optimization methods that dynamically adjust learning rate for each parameter, such as Adam~\cite{Adam} and AdaGrad~\cite{AdaGrad}. We empirically show that offline Gradient-Blending (Algorithm~\ref{algo:g_b_offline}) also works with different optimizers. Since SGD gives the best accuracy among the three optimizers, we opt to use SGD for all of our other experiments.

\begin{table}
    \captionsetup{font=footnotesize}
	\centering
	\small
	\begin{tabular}{|c|c|c|c|c|}
	\hline
	Optimizer & Method & Clip & V@1 & V@5 \\
	\hline
	\multirow{3}{4em}{AdaGrad} & Visual & 60.0 & 68.9 & 88.4 \\
	 & Naive AV & 56.4 & 65.2 & 86.5 \\
	 & G-Blend & \bf 62.1 & \bf 71.3 & \bf 89.8 \\
	\hline
	\multirow{3}{4em}{Adam} & Visual & 60.1 & 69.3 & 88.7 \\
	 & Naive AV & 57.9 & 66.4 & 86.8 \\
	 & G-Blend & \bf 63.0 & \bf 72.1 & \bf 90.5 \\
	\hline
	\end{tabular}
	\vspace{-3pt}
	\caption{\textbf{G-Blend on different optimizers.} We compare G-Blend with Visual only and Naive AV on two additional optimizers: AdaGrad, and Adam. G-Blend consistently outperforms Visual-Only and Naive AV baselines on all three optimizers.} 
	\label{tab:optimizer}
	\vspace{-6mm}
\end{table}

\noindent \textbf{Different Modalities.} On Kinetics, we study all combinations of three modalities: RGB, optical flow, and audio. Table~\ref{tab:different_modality_wl} presents comparison of our method with naive joint training and best single stream model. We observe significant gains of G-Blend compared to both baselines on all multi-modal problems. It is worth noting that G-Blend is generic enough to work for more than two modalities.

\begin{table*}[ht]
    \centering
    \captionsetup{font=footnotesize}
    {\small
    	\begin{tabularx}{\linewidth}{|c| *{12}{Y|}}
        	\hline
        	Modal & \multicolumn{3}{c|}{RGB + A} & \multicolumn{3}{c|}{RGB + OF} & \multicolumn{3}{c|}{OF + A} & \multicolumn{3}{c|}{RGB + OF + A}  \\
        	\hline
        	\footnotesize{Weights} & \multicolumn{3}{c|}{\scriptsize{[RGB,A,Join]=[0.630,0.014,0.356]}} & \multicolumn{3}{c|}{\scriptsize{[RGB,OF,Join]=[0.309,0.495,0.196]}} & \multicolumn{3}{c|}{\scriptsize{[OF,A,Join]=[0.827,0.011,0.162]}} & \multicolumn{3}{c|}{\scriptsize{[RGB,OF,A,Join]=[0.33,0.53,0.01,0.13]}} \\
        	\hline
        	Metric & Clip & V@1 & V@5 & Clip & V@1 & V@5 & Clip & V@1 & V@5 & Clip & V@1 & V@5\\
        	\hline
        	Uni & 63.5 & 72.6 & 90.1 & 63.5 & 72.6 & 90.1 & 49.2 & 62.1 & 82.6 & 63.5 & 72.6 & 90.1 \\
        	Naive & 61.8 & 71.4 & 89.3 & 62.2 & 71.3 & 89.6 & 46.2 & 58.3 & 79.9 & 61.0 & 70.0 & 88.7\\
        	\hline
        	\scriptsize{G-Blend} & \textbf{65.9} & \textbf{74.7} & \textbf{91.5} & \textbf{64.3} & \textbf{73.1} & \textbf{90.8} & \textbf{54.4} & \textbf{66.3} & \textbf{86.0} & \textbf{66.1} & \textbf{74.9} & \textbf{91.8} \\
        	\hline
    	\end{tabularx}
	}
	\vspace{-5pt}
	\caption{\textbf{Gradient-Blending (G-Blend) works on different multi-modal problems.} Comparison between G-Blend with naive late fusion and single best modality on Kinetics. On all 4 combinations of different modalities, G-Blend outperforms both naive late fusion network and best uni-modal network by large margins, and it also works for cases with more than two modalities. G-Blend results are averaged over three runs with different initialization. Variances are small and are provided in supplementary} 
	\label{tab:different_modality_wl}
	\vspace{-4mm} 
\end{table*}

\noindent \textbf{Different Architectures.} We conduct experiments on mid-fusion strategy~\cite{Owens_2018_ECCV}, which suffers less overfitting and outperforms visual baseline (Figure~\ref{fig:diff_approach_fail}). On audio-visual setting, Gradient-Blending gives 0.8\% improvement (top-1 from 72.8\% to 73.6\%). On a different fusion architecture with Low-Rank Multi-Modal Fusion (LMF)~\cite{LMF18}, Gradient-Blending gives 4.2\% improvement (top-1 from 69.3\% to 73.5\%). This suggests Gradiend-Blending can be adopted to other fusion strategies besides late-fusion and other fusion architectures besides concatenation. 

\noindent \textbf{Different Tasks/Benchmarks.} We pick the problem of joint audio-RGB model training, and go deeper to compare Gradient-Blending with other regularization methods on different tasks and benchmarks: action recognition (Kinetics), sport classification (mini-Sports), and acoustic event detection (mini-AudioSet). We include three baselines: adding dropout at concatenation layer~\cite{Dropout14}, pre-training single stream backbones then finetuning the fusion model, and blending the supervision signals with equal weights (which is equivalent to naive training with two auxiliary losses). Auxiliary losses are popularly used in multi-task learning, and we extend it as a baseline for multi-modal training.

As presented in Table \ref{tab:av_wl}, Gradient-Blending outperforms all baselines by significant margins on both Kinetics and mini-Sports. On mini-AudioSet, G-Blend improves all baselines on mAP, and is slightly worse on mAUC compared to auxiliary loss baseline. The reason is that the weights learned by Gradient-Blending are very similar to equal weights. The failures of auxiliary loss on Kinetics and mini-Sports demonstrates that the weights used in G-Blend are indeed important. We note that for mini-AudioSet, even though the naively trained multi-modal baseline is better than uni-modal baseline, Gradient-Blending still improves by finding more generalized information. We also experiment with other less obvious multi-task techniques such as treating the weights as learnable parameters~\cite{Kendall18}. However, this approach converges to a similar result as naive joint training. This happens because it lacks of overfitting prior, and thus the learnable weights were biased towards the head that has the lowest training loss which is audio-RGB. 

\begin{table*}
    \centering
    \captionsetup{font=footnotesize}
    \small
	\begin{tabularx}{\linewidth}{|c| *{8}{Y|}}
	\hline
	Dataset & \multicolumn{3}{c|}{Kinetics} & \multicolumn{3}{c|}{mini-Sports} & \multicolumn{2}{c|}{mini-AudioSet}\\
	\hline
	Weights & \multicolumn{3}{c|}{{[RGB,A,Join]=[0.63,0.01,0.36]}} & \multicolumn{3}{c|}{{[RGB,A,Join]=[0.65,0.06,0.29]}} & \multicolumn{2}{c|}{\scriptsize{[RGB,A,Join]=[0.38,0.24,0.38]}}\\
	\hline
	Method & Clip & V@1 & V@5 & Clip & V@1 & V@5 & mAP & mAUC\\
	\hline
	Audio only &  13.9 & 19.7 & 33.6 & 14.7 & 22.1 & 35.6 & 29.1 & 90.4\\
	RGB only & 63.5 & 72.6 & 90.1 & 48.5 & 62.7 & 84.8 & 22.1 & 86.1\\
	\hline
	Pre-Training & 61.9 & 71.7 & 89.6 & 48.3 & 61.3 & 84.9 & 37.4 & 91.7\\
	Naive & 61.8 & 71.7 & 89.3 & 47.1 & 60.2 & 83.3 & 36.5 & 92.2\\
	Dropout & 63.8 & 72.9 & 90.6 & 47.4 & 61.4 & 84.3 & 36.7 & 92.3\\
	Auxiliary Loss & 60.5 & 70.8 & 88.6 & 48.9 & 62.1 & 84.0 & 37.7 & \textbf{92.3}\\
	\hline
	G-Blend & \textbf{65.9} & \textbf{74.7} & \textbf{91.5} & \textbf{49.7} & \textbf{62.8} & \textbf{85.5} & \textbf{37.8} & 92.2\\
	\hline
	\end{tabularx} 
	\vspace{-5pt}
	\caption{\textbf{G-Blend outperforms all baseline methods on different benchmarks and tasks.} Comparison of G-blend with different regularization baselines as well as uni-modal networks on Kinetics, mini-Sports, and mini-AudioSet. G-Blend consistently outperforms other methods, except for being comparable with using auxiliary loss on mini-AudioSet due to the similarity of learned weights of G-Blend and equal weights.} 
	\label{tab:av_wl}
	\vspace{-5mm}
\end{table*}

\begin{figure}[h!]
 \centering
 \captionsetup{font=footnotesize}
 \begin{subfigure}[b]{0.43\linewidth}
    {\includegraphics[width=\linewidth]{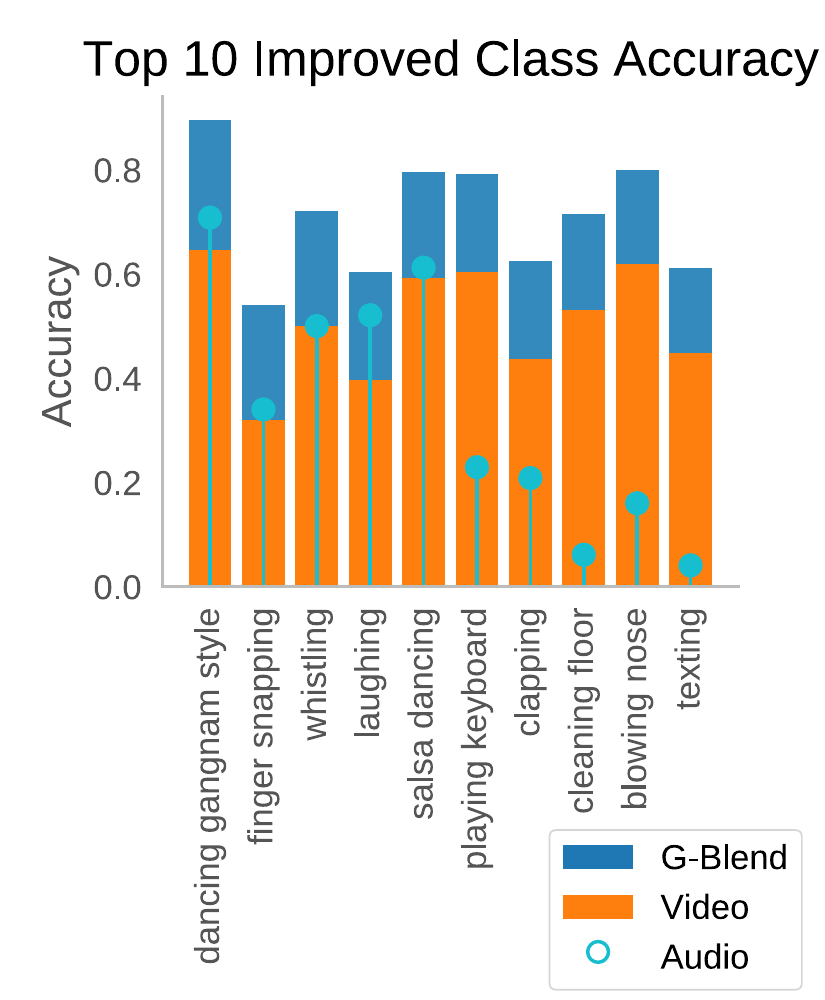}}
 \end{subfigure}
 \begin{subfigure}[b]{0.43\linewidth}
    {\includegraphics[width=\linewidth]{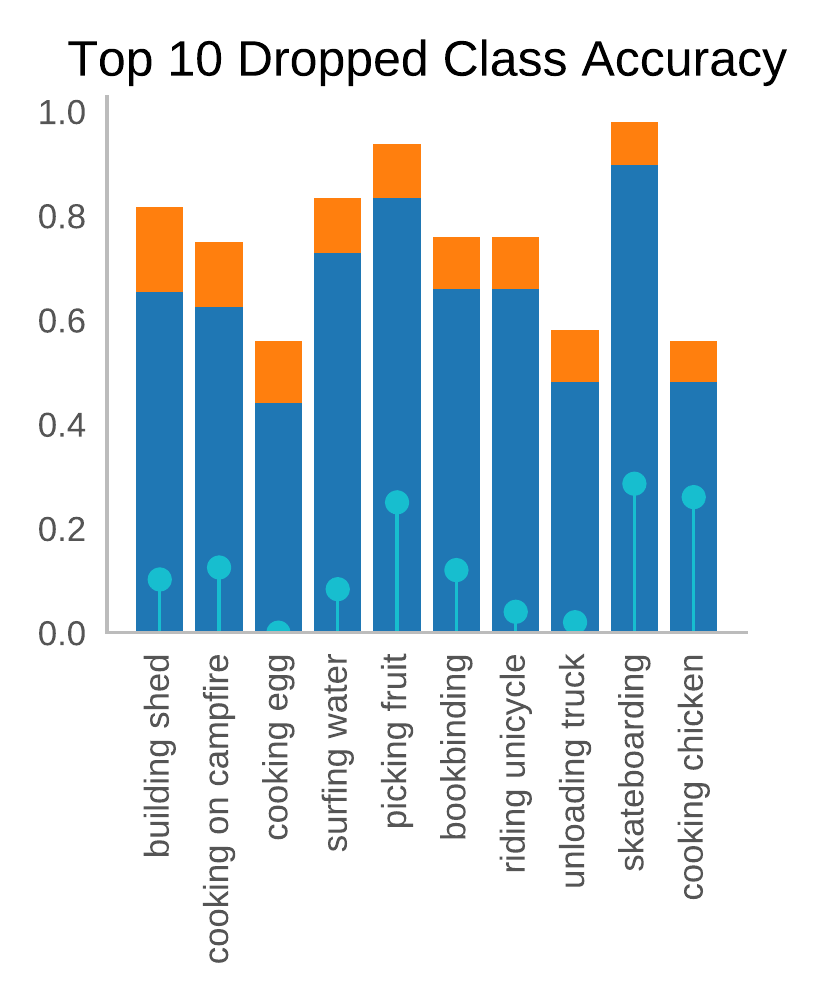}}
 \end{subfigure}
 \vspace{-3mm} 
 \caption{\textbf{Top-Bottom 10 classes based on improvement of G-Blend to RGB model.} The improved classes are indeed audio-relevant, while those have performance drop are not very audio semantically-related.}
 \vspace{-3mm} 
 \label{fig:top_bot_10}
\end{figure}

Fig.~\ref{fig:top_bot_10} presents top and bottom 10 classes on Kinetics where G-Blend makes the most and least improvements compared with RGB-only. We observe that improved classes usually have a strong audio-correlation, such as clapping and laughing. For texting, although audio-only has nearly 0 accuracy, when combined with RGB using G-Blend, there are still significant improvements. On bottom-10 classes, we indeed find that audio does not seem to be very semantically relevant (e.g. unloading truck). See supplementary materials for more qualitative analysis.

\section{Comparison with State-of-the-Art}
\label{sec:compare_with_sota}

In this section, we train our multi-modal networks with deeper backbone architectures using offline Gradient-Blending and compare them with state-of-the-art methods on Kinetics, EPIC-Kitchen~\cite{Damen2018EPICKITCHENS}, and AudioSet. EPIC-Kitchen is a multi-class egocentric dataset with ~28K training videos associated with 352 noun and 125 verb classes. For ablations, following~\cite{Baradel2018ObjectReason}, we construct a validation set of unseen kitchen environments. G-Blend is trained with RGB and audio input. For Kinetics and EPIC-Kitchen, we use ip-CSN~\cite{CSN19} for visual backbone with 32 frames and ResNet for audio backbone, both with 152 layers. For AudioSet, we use R(2+1)D for visual~\cite{Tran18} with 16 frames and ResNet for audio, both with 101 layers. We use the same training setup in section~\ref{sec:ablation}. For EPIC-Kitchen, we follow the same audio feature extractions as~\cite{EPIC-FUSION}; the visual backbone is pre-trained on IG-65M~\cite{Ghadiyaram2019LargescaleWP}. We use the same evaluation setup as section~\ref{sec:ablation} for AudioSet and EPIC-Kitchen. For Kinetics, we follow the 30-crop evaluation setup as~\cite{XiaolongWang18}. Our main purposes in these experiments are: 1) to confirm the benefit of Gradient-Blending on high-capacity models; and 2) to compare G-Blend with state-of-the-art methods on different large-scale benchmarks.


\begin{table}[]
    \centering
    \small 
    \captionsetup{font=footnotesize}
   \begin{tabular}{|c|c|c|c|c|}
    	\hline
    	\bf Backbone & {Pre-train} & V@1 & V@5 & GFLOPs \\
    	\hline
    	\scriptsize{Shift-Attn Net \cite{abs-1708-03805}} & ImageNet & 77.7 & 93.2 & NA \\
    	\scriptsize{SlowFast \cite{slowfast}} & None & 78.9 & 93.5 & 213$\times$30 \\
    	\scriptsize{SlowFast+NL \cite{slowfast}} & None & 79.8 & 93.9 & 234$\times$30 \\
    	\hline
    	\scriptsize{ip-CSN-152 \cite{CSN19}} & None & 77.8 & 92.8 & {108.8$\times$30} \\ 
    	\footnotesize{\bf G-Blend(ours)} & None & 79.1 & 93.9 & {110.1$\times$30} \\
    	\hline
    	\scriptsize{ip-CSN-152 \cite{CSN19}} & {Sports1M} & 79.2 & 93.8 & {108.8$\times$30} \\ 
    	\footnotesize{\bf G-Blend(ours)} & {Sports1M} & \bf 80.4 & \bf 94.8 & {110.1$\times$30} \\
    	\hline 
    	\scriptsize{ip-CSN-152 \cite{CSN19}} & IG-65M & 82.5 & 95.3 & {108.8$\times$30} \\ 
    	\footnotesize{\bf G-Blend(ours)} & IG-65M & \bf 83.3 & \bf 96.0 & {110.1$\times$30} \\
    	\hline
    	\end{tabular}
    	\vspace{-5pt}
    	\caption{{\bf Comparison with state-of-the-art methods on Kinetics.} G-Blend used audio and RGB as input modalities; for pre-trained models on Sports1M and IG-65M, G-Blend initializes audio network by pre-training on AudioSet. G-Blend outperforms current state-of-the-art multi-modal method (Shift-Attention Network) despite the fact that it uses fewer modalities (G-Blend does not use Optical Flow). G-Blend also gives a good improvement over RGB model (the best uni-modal network) when using the same backbone, and it achieves the state-of-the-arts.} 
    \label{tab:sota_kinetics}
    \vspace{-5mm}
\end{table}

\noindent {\bf Results.} Table~\ref{tab:sota_kinetics} presents results of G-Blend and compares them with current state-of-the-art methods on Kinetics. First, G-Blend provides an 1.3\% improvement over RGB model (the best uni-modal network) with the same backbone architecture ip-CSN-152~\cite{CSN19} when both models are trained from scratch. This confirms that the benefits of G-Blend still hold with high capacity model. Second, G-Blend outperforms state-of-the-arts multi-modal baseline Shift-Attention Network~\cite{abs-1708-03805} by 1.4\% while using less modalities (not using optical flow) and no pre-training. It is on-par with SlowFast~\cite{slowfast} while being 2x faster. G-Blend, when fine-tuned from Sports-1M on visual and AudioSet on audio, outperforms SlowFast Network and SlowFast augmented by Non-Local~\cite{XiaolongWang18} by 1.5\% and 0.6\% respectively, while being 2x faster than both. Using weakly-supervised pre-training by IG-65M~\cite{Ghadiyaram2019LargescaleWP} on visual, G-Blend gives unparalleled 83.3\% top-1 accuracy and 96.0\% top-5 accuracy. 

We also note that there are many competitive methods reporting results on Kinetics, due to the space limit, we select only a few representative methods for comparison including Shift-Attention~\cite{abs-1708-03805}, SlowFast~\cite{slowfast}, and ip-CSN~\cite{CSN19}. Shift-Attention and SlowFast are the methods with the best published accuracy using multi-modal and uni-modal input, respectively. ip-CSN is used as the visual backbone of G-Blend thus serves as a direct baseline. 



\begin{table}[]
    \centering
    \small 
    \captionsetup{font=footnotesize}
        \begin{tabular}{|c|c|c|}
    	\hline
    	{\bf Method} & mAP & mAUC \\
    	\hline
    	Multi-level Attn. \cite{YuMultilvl18} & 0.360 & 0.970 \\
    	TAL-Net \cite{TalNet} & 0.362 & 0.965 \\
    	\hline
    	Audio:R2D-101 & 0.324 & 0.961 \\
    	Visual:R(2+1)D-101 & 0.188 & 0.918 \\
    	Naive A/V:101 & 0.402 & 0.973 \\
    	\hline
    	{\bf G-Blend (ours)}:101 & \textbf{0.418} & \textbf{0.975} \\
    	\hline
    	\end{tabular} 
    	\vspace{-5pt}
    	\caption{{\bf Comparison with state-of-the-art methods on AudioSet.} G-Blend outperforms the state-of-the-art methods by a large margin.}
    \label{tab:sota_audioset}
    \vspace{-5mm}
\end{table}

\begin{table}[]
    \centering
    \footnotesize
    \captionsetup{font=footnotesize}
        \begin{tabular}{|c|c|c|c|c|c|c|}
        \hline
        method & \multicolumn{2}{c|}{noun} & \multicolumn{2}{c|}{verb} & \multicolumn{2}{c|}{action} \\
        \hline
         & \scriptsize V@1 & \scriptsize V@5 & \scriptsize V@1 & \scriptsize V@5 & \scriptsize V@1 & \scriptsize V@5 \\
    	\hhline{|=======|}
    	\multicolumn{7}{|c|}{Validation Set} \\
    	\hline
    	\scriptsize{Visual:ip-CSN-152~\cite{CSN19}} & \bf 36.4 & \bf 58.9 & 56.6 & 84.1 & 24.9 & 42.5 \\
    	\scriptsize Naive A/V:152 & 34.8 & 56.7 & 57.4 & 83.3 & 23.7 & 41.2 \\
    	\hline 
    	\scriptsize G-Blend(ours) & \underline{36.1} & \underline{58.5} & \bf 59.2 & \bf 84.5 & \bf 25.6 & \bf 43.5 \\
    	\hhline{|=======|}
    	\multicolumn{7}{|c|}{Test Unseen Kitchen (S2)} \\
    	\hline
    	\scriptsize Leaderboard~\cite{Epic-Challenge} & \bf 38.1 & \bf 63.8 & \bf 60.0 & \underline{82.0} & \bf 27.4 & \underline{45.2} \\
    	\hline
    	\scriptsize Baidu-UTS~\cite{BaiduUTSKitchen} & 34.1 & \underline{62.4} & \bf 59.7 & \bf 82.7 & 25.1 & \bf 46.0 \\
    	\scriptsize TBN Single~\cite{EPIC-FUSION} & 27.9 & 53.8 & 52.7 & 79.9 & 19.1 & 36.5 \\
    	\scriptsize TBN Ensemble~\cite{EPIC-FUSION} & 30.4 & 55.7 & 54.5 & {81.2} & 21.0 & 39.4 \\
    	\scriptsize Visual:ip-CSN-152 & 35.8 & 59.6 & 56.2 & 80.9 & 25.1 & 41.2 \\
    	\hline
    	\scriptsize G-Blend(ours) & \underline{36.7} & 60.3 & 58.3 & 81.3 & \underline{26.6} &  43.6 \\
    	\hhline{|=======|}
    	\multicolumn{7}{|c|}{Test Seen Kitchen (S1)} \\
    	\hline
    	\scriptsize Baidu-UTS(leaderboard)  & {\bf 52.3} & \bf 76.7 & \bf 69.8 & \bf 91.0 & \bf 41.4 & \bf 63.6 \\
    	\scriptsize TBN Single & 46.0 & 71.3 & 64.8 & 90.7 & 34.8 & 56.7 \\
    	\scriptsize TBN Ensemble & 47.9 & \underline{72.8} & 66.1 & \bf 91.2 & 36.7 & \underline{58.6} \\
    	\scriptsize Visual:ip-CSN-152 & 45.1 & 68.4 & 64.5 & 88.1 & 34.4 & 52.7 \\
    	\hline
    	\scriptsize G-Blend(ours) & \underline{48.5} & 71.4 & \underline{66.7} & 88.9 & \underline{37.1} & 56.2 \\
    	\hline
    	\end{tabular} 
    	\vspace{-5pt}
    	\caption{{\bf Comparison with state-of-the-art methods on EPIC-Kitchen.} G-Blend achieves 2nd place on seen kitchen challenge and 4th place on unseen, despite using fewer modalities, fewer backbones, and single model in contrast to model ensembles compared to published results on leaderboard.}
    \label{tab:sota_kitchen}
    \vspace{-5mm}
\end{table}

Table~\ref{tab:sota_audioset} presents G-Blend results on AudioSet. Since AudioSet is very large (2M), we use mini-AudioSet to estimate weights. G-Blend outperforms two state-of-the-art Multi-level Attention Network\cite{YuMultilvl18} and TAL-Net\cite{TalNet} by 5.8\% and 5.5 \% on mAP respectively, although the first one uses strong features (pre-trained on YouTube100M) and the second uses 100 clips per video, while G-Blend uses only 10. 

Table~\ref{tab:sota_kitchen} presents G-Blend results and compare with published SoTA results and leaderboard on the EPIC-Kitchens Action Recognition challenge. On validation set, G-Blend outperforms naive A/V baseline on noun, verb and action; it is on par with visual baseline on noun and outperforms visual baseline on verb and action. Currently, G-Blend ranks the 2nd place on unseen kitchen in the challenge and 4th place on seen kitchen. Comparing to published results, G-Blend uses less modalities (not using optical flow as TBN Ensemble~\cite{EPIC-FUSION}), less backbones (Baidu-UTS~\cite{BaiduUTSKitchen} uses three 3D-CNNs plus two detection models), and a single model (TBN Ensemble~\cite{EPIC-FUSION} uses ensemble of five models).

\section{Discussion}
\label{sec:conclusion}

In uni-modal networks, diagnosing and correcting overfitting typically involves manual inspection of learning curves. Here we have shown that for multi-modal networks it is essential to measure and correct overfitting in a principled way, and we put forth a useful and practical measure of overfitting. Our proposed method, Gradient-Blending, uses this measure to obtain significant improvements over baselines, and either outperforms or is comparable with state-of-the-art methods on multiple tasks and benchmarks. The method potentially applies broadly to end-to-end training of ensemble models, and we look forward to extending G-Blend to other fields where calibrating multiple losses is needed, such as multi-task.




\newpage

{
\bibliographystyle{ieee}
\bibliography{ieeedu_ref}
}

\newpage
\appendix

\section{Estimating Weights on Subsets of Data}

\begin{figure}
\captionsetup{font=footnotesize}
\begin{center}
\includegraphics[width=0.7\linewidth]{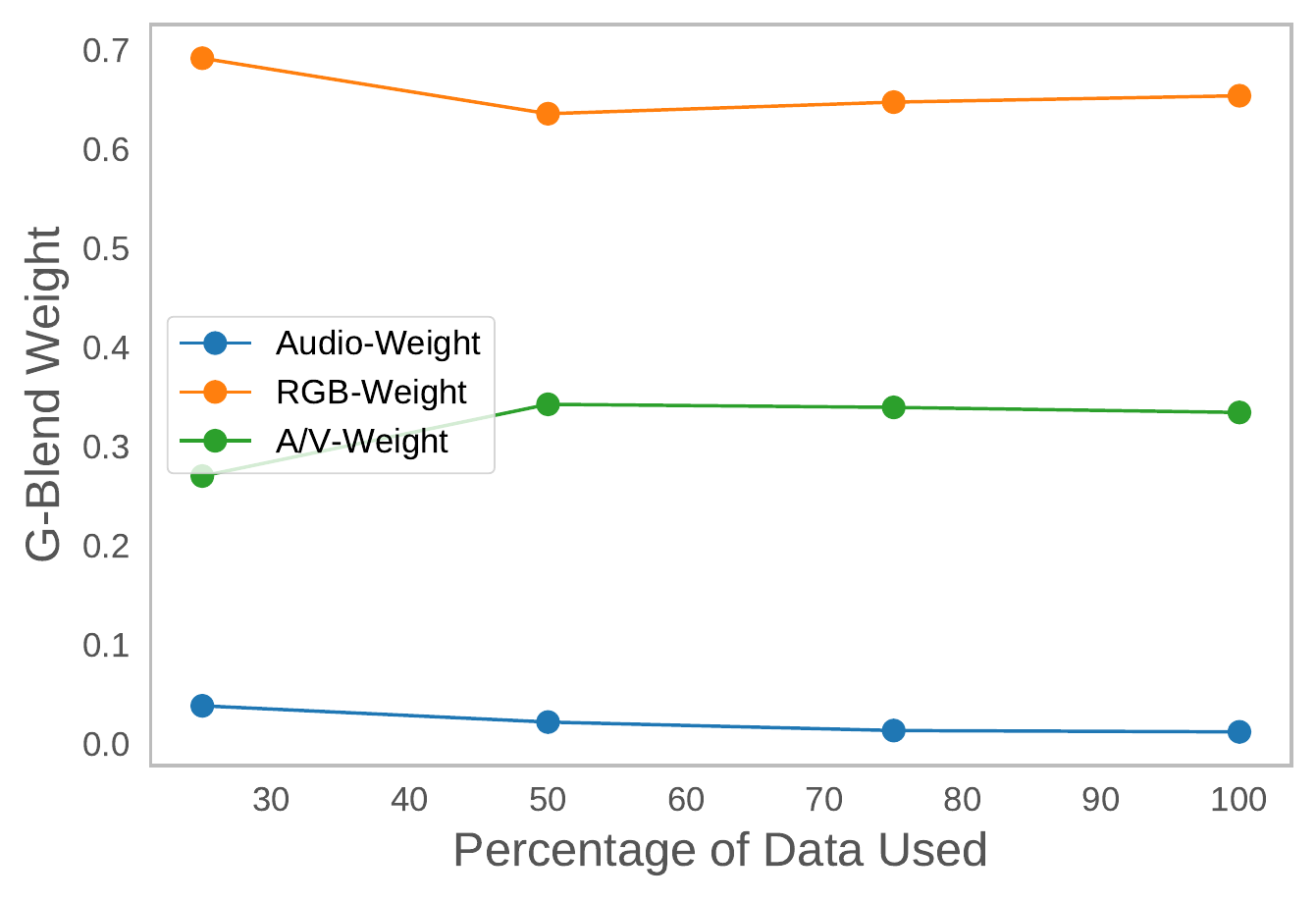}
\caption{{\bf Weight Estimations on Subsets of Data}. We used a small subset of Kinetics dataset to estimate the weights. The weights are quite robust as we decrease the volume of dataset. This suggests feasibility to use subsets to reduce the costs for Gradient-Blending.}
\label{fig:subset_weights}
\end{center}
\vspace{-5mm} 
\end{figure}

We show that weight estimations by Gradient-Blending is robust on small subsets of data. We sampled 25\%, 50\% and 75\% of Kinetics dataset and use these subsets as train sets in Alg.~2 in main paper. As shown in Fig.~\ref{fig:subset_weights}, the estimated weights are stable on small subsets of data. This suggests that the computational cost of the algorithm can be reduced by using a small subset of data for weight estimation. 

\section{Understanding $OGR$}

Overfitting is typically understood as learning patterns in a training set that do not generalize to the target distribution. We quantify this as follows. Given model parameters $\Theta^{(N)}$, where $N$ indicates the training epoch, let $\mathcal{L}^{\mathcal{T}}(\Theta^{(N)})$ be the model's average loss over the fixed training set, and $\mathcal{L}^*(\Theta^{(N)})$ be the ``true'' loss w.r.t the hypothetical target distribution. (In practice, $\mathcal{L}^*$ is approximated by the test and validation losses.) For either loss, the quantity $\mathcal{L}(\Theta^{(0)}) - \mathcal{L}(\Theta^{(N)})$ is a measure of the information gained during training. We define overfitting as the gap between the gain on the training set and the target distribution:
\begin{align*}
    O_N \equiv \left( \mathcal{L}^\mathcal{T}(\Theta^{(0)}) - \mathcal{L}^\mathcal{T}(\Theta^{(N)}) \right) - \left( \mathcal{L}^*(\Theta^{(0)}) - \mathcal{L}^*(\Theta^{(N)}) \right)
\end{align*}
and generalization to be the amount we learn (from training) about the target distribution:
\begin{align*}
    G_N \equiv \mathcal{L}^*(\Theta^{(0)}) - \mathcal{L}^*(\Theta^{(N)})
\end{align*}
The overfitting-to-generalization ratio is a measure of information quality for the training process of $N$ epochs:
\begin{equation}
    \resizebox{0.9\hsize}{!}{$OGR = \left| \frac{\left( \mathcal{L}^\mathcal{T}(\Theta^{(0)}) - \mathcal{L}^\mathcal{T}(\Theta^{(N)})\right) - \left( \mathcal{L}^*(\Theta^{(0)}) - \mathcal{L}^*(\Theta^{(N)})\right)}{\mathcal{L}^*(\Theta^{(0)}) - \mathcal{L}^*(\Theta^{(N)})} \right|$}
\label{equ:OLR_rewrite}
\end{equation}

We can also define the amount of overfitting and generalization for an intermediate step from epoch $N$ to epoch $N+n$, where 
\begin{align*}
    \Delta O_{N,n} \equiv \left( O_{N+n} - O_N \right)
\end{align*}
and 
\begin{align*}
    \Delta G_{N,n} \equiv \left( G_{N+n} - G_N \right)
\end{align*}

Together, this gives $OGR$ between any two checkpoints:

\begin{align*}
    OGR \equiv \langle \frac{\Delta O_{N,n}}{\Delta G_{N,n}} \rangle
\end{align*}

However, it does not make sense to optimize this as-is. Very underfit models, for example, may still score quite well (difference of train loss and validation loss is very small for underfitting models).  What does make sense, however, is to solve an infinitesimal problem: given several estimates of the gradient, blend them to minimize an infinitesimal $OGR$ (or equivalently $OGR^2$).  We can then apply this blend to our optimization process by stochastic gradients (eg. SGD with momentum).  In a multi-modal setting, this means we can combine gradient estimates from multiple modalities and minimize $OGR$ to ensure each gradient step now produces a gain no worse than that of the single best modality.






Consider this in an infinitesimal setting (or a single parameter update step). Given parameter $\Theta$, the full-batch gradient with respect to the training set is $\nabla \mathcal{L}^{\mathcal{T}}(\Theta)$, and the groundtruth gradient is $\nabla \mathcal{L}^*(\Theta)$. We decompose $\nabla \mathcal{L}^{\mathcal{T}}$ into the true gradient and a remainder:
\begin{equation}
    \nabla \mathcal{L}^{\mathcal{T}}(\Theta) = \nabla \mathcal{L}^*(\Theta) + \epsilon
\end{equation}
In particular, $\epsilon = \nabla \mathcal{L}^{\mathcal{T}}(\Theta) - \nabla \mathcal{L}^*(\Theta)$ is exactly the infinitesimal overfitting. Given an estimate $\hat{g}$ with learning rate $\eta$, we can measure its contribution to the losses via Taylor's theorem:
\begin{align*}
    \mathcal{L}^{\mathcal{T}}(\Theta + \eta \hat{g}) \approx \mathcal{L}^{\mathcal{T}}(\Theta) + \eta \langle \nabla \mathcal{L}^{\mathcal{T}}, \hat{g} \rangle \\
    \mathcal{L}^{*}(\Theta + \eta \hat{g}) \approx \mathcal{L}^{*}(\Theta) + \eta \langle \nabla \mathcal{L}^{*}, \hat{g} \rangle
\end{align*}
which implies $\hat{g}$'s contribution to overfitting is given by $\langle \nabla \mathcal{L}^{\mathcal{T}} - \nabla \mathcal{L}^{*}, \hat{g} \rangle$.  
If we train for $N$ steps with gradients $\{\hat{g}_i\}_{0}^{N}$, and $\eta_i$ is the learning rate at $i$-th step, the final $OGR$ can be aggregated as:
\begin{equation}
    OGR = \left| \frac{\sum_{i=0}^N \eta_i \langle \nabla \mathcal{L}^{\mathcal{T}}(\Theta^{(i)}) - \nabla \mathcal{L}^*(\Theta^{(i)}), \hat{g}_i \rangle}{\sum_{i=0}^N \eta_i \langle \nabla \mathcal{L}^*(\Theta^{(ni}), \hat{g}_i \rangle} \right| 
\label{equ:OLR_agg}
\end{equation}
and $OGR^2$ for a single vector $\hat{g}_i$ is
\begin{equation}
    OGR^2 = \left( \frac{ \langle \nabla \mathcal{L}^{\mathcal{T}}(\Theta^{(i)}) - \nabla \mathcal{L}^*(\Theta^{(i)}), \hat{g}_i \rangle}{\langle \nabla \mathcal{L}^*(\Theta^{(i)}), \hat{g}_i \rangle} \right)^2 
\label{equ:OLR}
\end{equation}
Next we will compute the optimal blend to minimize single-step $OGR^2$.




\section{Proof of Proposition~1} \label{supplement_proof}

\begin{proof}[Proof of Proposition~1]
Without loss of generality, we solve the problem with a different normalization:
\begin{align}
\langle \nabla \mathcal{L}^* , \sum_k w_k v_k \rangle =1
\label{equ:generalization_constraint}
\end{align}
(Note that one can pass between normalizations simply by uniformly rescaling the weights.)  With this constraint, the problem simplifies to:

\begin{equation}
    w^* = \argmin_{w} \E[(\langle \nabla \mathcal{L}^{\mathcal{T}} - \nabla \mathcal{L}^*,\sum_k w_k v_k \rangle)^2]
\end{equation}

We first compute the expectation:

\begin{align}
    & \E[(\langle \nabla \mathcal{L}^{\mathcal{T}} - \nabla \mathcal{L}^*,\sum_k w_k v_k \rangle)^2] \nonumber \\
    &= \E[(\sum_k w_k \langle \nabla \mathcal{L}^{\mathcal{T}} - \nabla \mathcal{L}^*,v_k \rangle)^2] \nonumber \\
    &= \E[\sum_{k,j}w_k w_j  \langle \nabla \mathcal{L}^{\mathcal{T}} - \nabla \mathcal{L}^*, v_k\rangle \langle \nabla \mathcal{L}^{\mathcal{T}} - \nabla \mathcal{L}^*, v_j\rangle ] \nonumber \\
    &= \sum_{k,j}w_k w_j \E \left[ \langle \nabla \mathcal{L}^{\mathcal{T}} - \nabla \mathcal{L}^*, v_k\rangle \langle \nabla \mathcal{L}^{\mathcal{T}} - \nabla \mathcal{L}^*, v_j\rangle \right] \nonumber \\
    &= \sum_k w_k^2 \sigma_k^2 \label{equ:approx_exp}
\end{align}

where $\sigma_k^2 = \E[\langle \nabla \mathcal{L}^{\mathcal{T}} - \nabla \mathcal{L}^*, v_k \rangle^2]$ and the cross terms vanish by assumption. 

We apply Lagrange multipliers on our objective function (\ref{equ:approx_exp}) and constraint (\ref{equ:generalization_constraint}):
\begin{equation}
    \mathit{L} = \sum_k w_k^2 \sigma_k^2 - \lambda\left(\sum_k w_k \langle \nabla\mathcal{L}^*, v_k \rangle - 1\right)
\end{equation}
The partials with respect to $w_k$ are given by 
\begin{equation}
    \frac{\partial \mathit{L}}{\partial w_k} = 2w_k \sigma_k^2 - \lambda \langle \nabla \mathcal{L}^*, v_k \rangle
\end{equation}
Setting the partials to zero, we obtain the weights:
\begin{equation}
    w_k = \lambda \frac{\langle \nabla \mathcal{L}^*, v_k \rangle}{2\sigma_k^2}
\end{equation}
The only remaining task is obtaining the normalizing constant.  Applying the constraint gives:
\begin{equation}
    1 = \sum_k w_k \langle \nabla \mathcal{L}^*, v_k \rangle = \lambda \sum_k \frac{\langle \nabla \mathcal{L}^*, v_k \rangle^2}{2\sigma_k^2}
\end{equation}
In other words,
\begin{equation}
    \lambda = \frac{2}{\sum_k \frac{\langle \nabla \mathcal{L}^*, v_k \rangle^2}{\sigma_k^2}}
\end{equation}

Setting $Z = 1 / \lambda$ we obtain $w_k^* = \frac{1}{Z}\frac{\langle \nabla \mathcal{L}^*, v_k \rangle^2}{2\sigma_k^2}$.  Dividing by the sum of the weights yields the original normalization.

\end{proof}

Note: if we relax the assumption that $\E[\langle \nabla \mathcal{L}^{\mathcal{T}} - \nabla \mathcal{L}^*, v_k\rangle \langle \nabla \mathcal{L}^{\mathcal{T}} - \nabla \mathcal{L}^*, v_j\rangle] = 0$ for $k \ne j$, the proof proceeds similarly, although from (\ref{equ:approx_exp}) it becomes more convenient to proceed in matrix notation.  Define a matrix $\Sigma$ with entries given by
\begin{align*}
    \Sigma_{kj} = \E[\langle \nabla \mathcal{L}^{\mathcal{T}} - \nabla \mathcal{L}^*, v_k\rangle \langle \nabla \mathcal{L}^{\mathcal{T}} - \nabla \mathcal{L}^*, v_j\rangle]
\end{align*}
Then one finds that 
\begin{align*}
    w_k^* &= \frac{1}{Z} \sum_j \Sigma^{-1}_{kj} \langle \nabla \mathcal{L}^*, v_k\rangle \\
    Z &= \frac{1}{2} \sum_{k,j} \Sigma^{-1}_{kj} \langle \nabla \mathcal{L}^*, v_k\rangle^2
\end{align*}

\section{Variances of G-Blend Runs}

\begin{table*}[!h]
    \centering
    \captionsetup{font=footnotesize}
    {\small
    	\begin{tabular}{|c|c|c|c|c|c|c|c|c|c|c|c|}
        	\hline
        	\multicolumn{3}{|c|}{RGB + A} & \multicolumn{3}{c|}{RGB + OF} & \multicolumn{3}{c|}{OF + A} & \multicolumn{3}{c|}{RGB + OF + A}  \\
        	\hline
        	Clip & V@1 & V@5 & Clip & V@1 & V@5 & Clip & V@1 & V@5 & Clip & V@1 & V@5\\
        	\hline
        	\scriptsize\textbf{65.9$\pm$0.1} & \scriptsize\textbf{74.7$\pm$0.2} & \scriptsize\textbf{91.5$\pm$0.1} & \scriptsize\textbf{64.3$\pm$0.1} & \scriptsize\textbf{73.1$\pm$0.0} & \scriptsize\textbf{90.8$\pm$0.1} & \scriptsize\textbf{54.4$\pm$0.6} & \scriptsize\textbf{66.3$\pm$0.5} & \scriptsize\textbf{86.0$\pm$0.6} & \scriptsize\textbf{66.1$\pm$0.4} & \scriptsize\textbf{74.9$\pm$0.2} & \scriptsize\textbf{91.8$\pm$0.2} \\
        	\hline
    	\end{tabular}
	}
	\vspace{3pt}
	\caption{\textbf{Last row of Table~3 in main papers with variance.} Results are averaged over three runs with random initialization, and $\pm$ indicates variances.} 
	\label{tab:small_var}
\end{table*}

The variances of the performances on the datasets used by the paper are typically small, and previous works provide results on a single run. To verify that G-Blend results are reproducible, we conducted multiple runs for G-Blend results in Table~3 of the main paper. We found that the variance is consistent across different modalities for G-Blend results (Table~\ref{tab:small_var}).

\section{Sub-sampling and Balancing Multi-label Dataset} \label{supp:bal_audioset}

For a single-label dataset, one can subsample and balance at a per-class level such that each class may have the same volume of data. Unlike single-label dataset, classes in multi-label dataset can be correlated. As a result, sampling a single data may add volume for more than one class. This makes the naive per-class subsampling approach difficult. 

To uniformly sub-sample and balance AudioSet to get mini-AudioSet, we propose the following algorithm:

\begin{algorithm}
\SetAlgoLined
\KwData{Original Multi-Class Dataset $\mathcal{D}$, Minimum Class Threshold $M$, Target Class Volume $N$}
\KwResult{Balanced Sub-sampled Multi-label Dataset $\mathcal{D}'$}
 Initialize empty dataset $\mathcal{D}'$ \;
 Remove labels from $\mathcal{D}$ such that label volume is less than $M$\;
 Randomly shuffle entries in $\mathcal{D}$\;
 \For{Data Entry $d\in \mathcal{D}$}{
  Choose class $c$ of $d$ such that the volume of $c$ is the smallest in $\mathcal{D}'$ \;
  Let the volume of $c$ be $V_c$ in $\mathcal{D}$ \;
  Let the volume of $c$ be ${V_c}'$ in $\mathcal{D}'$ \;
  Generate random number $r$ to be an integer between $0$ and $V_c - {V_c}'$ \;
  \eIf{$r < N - {V_c}'$}{
   Select $d$ to $\mathcal{D}'$ \;
   }{
   Skip $d$ and continue \;
  }
 }
 \caption{Sub-sampling and Balancing Multi-label Dataset}
\label{algo:bal_audioset}
\end{algorithm}

\section{Details on Model Architectures}

\subsection{Late Fusion By Concatenation}

In late fusion by concatenation strategy, we concatenate the output features from each individual network (i.e. $k$ modalities' 1-D vectors with $n$ dimensions). If needed, we add dropout after the feature concatenations. 

The fusion network is composed of two \emph{FC} layers, with each followed by an \emph{ReLU} layer, and a linear classifier. The first \emph{FC} maps $kn$ dimensions to $n$ dimensions, and the second one maps $n$ to $n$. The classifier maps $n$ to $c$, where $c$ is the number of classes. 

As sanity check, we experimented using less or more $FC$ layers on Kinetics:
\begin{itemize}[noitemsep]
    \item \textbf{0 \emph{FC}.} We only add a classifier that maps $kn$ dimensions to $c$ dimensions. 
    \item \textbf{1 \emph{FC}.} We add one \emph{FC} layer that maps $kn$ dimensions to $n$ dimension, followed by an \emph{ReLU} layer and classifier to map $n$ dimension to $c$ dimensions.
    \item \textbf{4 \emph{FC}.} We add one \emph{FC} layer that maps $kn$ dimensions to $n$ dimension, followed by an \emph{ReLU} layer. Then we add 3 $\emph{FC-ReLU}$ pairs that preserve the dimensions. Then we add an a classifier to map $n$ dimension to $c$ dimensions.
\end{itemize}

We noticed that the results of all these approaches are sub-optimal. We speculate that less layers may fail to fully learn the relations of the features, while deeper fusion network overfits more.

\subsection{Mid Fusion By concatenation}

Inspired by \cite{Owens_2018_ECCV}, we also concatenate the features from each stream at an early stage rather than late fusion. The problem with mid fusion is that features from individual streams can have different dimensions. For example, audio features are 2-D (time-frequency) while visual features are 3-D (time-height-width). 

We propose three ways to match the dimension, depending on the output dimension of the concatenated features:

\begin{itemize}[noitemsep]
    \item \textbf{1-D Concat.} We downsample the audio features to 1-D by average pooling on the frequency dimension. We downsample the visual features to 1-D by average pooling over the two spatial dimensions. 
    \item \textbf{2-D Concat.} We keep the audio features the same and match the visual features to audio features. We downsample the visual features to 1-D by average pooling over the two spatial dimensions. Then we tile the 1-D visual features on frequency dimension to make 2-D visual features.
    \item \textbf{3-D Concat.} We keep the visual features fixed and match the audio features to visual features.  We downsample the audio features to 1-D by average pooling over the frequency dimension. Then we tile the 1-D visual features on two spatial dimensions to make 3-D features.
\end{itemize}

The temporal dimension may also be mismatched between the streams: audio stream is usually longer than visual streams. We add convolution layers with stride of 2 to downsample audio stream if we are performing 2-D concat. Otherwise, we upsample visual stream by replicating features on the temporal dimension.

There are five blocks in the backbones of our ablation experiments (section 4), and we fuse the features using all three strategies after block 2, block 3, and block 4. Due to memory issue, fusion using 3-D concat after block 2 is unfeasible. On Kinetics, we found 3-D concat after block 3 works the best, and it's reported in Fig.~1 in the main paper. In addition, we found 2-D concat works the best on AudioSet and uses less GFLOPs than 3-D concat. We speculate that the method for dimension matching is task-dependent.


\subsection{SE Gate}

Squeeze-and-Excitement network introduced in \cite{SENet} applies a self-gating mechanism to produce a collection of per-channel weights. Similar strategies can be applied in a multi-modal network to take inputs from one stream and produce channel weights for the other stream. 

Specifically, we perform global average pooling on one stream and use the same architectures in \cite{SENet} to produce a set of weights for the other channel. Then we scale the channels of the other stream using the weights learned. We either do a ResNet-style skip connection to add the new features or directly replace the features with the scaled features. The gate can be applied from one direction to another, or on both directions. The gate can also be added at different levels for multiple times. We found that on Kinetics, it works the best when applied after block 3 and on both directions. 

We note that we can also first concatenate the features and use features from both streams to learn the per-channel weights. The results are similar to learning the weights with a single stream. 

\subsection{NL Gate}

Although lightweight, SE-gate fails to offer any spatial-temporal or frequency-temporal level attention. One alternative way is to apply an attention-based gate. We are inspired by the Query-Key-Value formulation of gates in \cite{AttentionAll17}. For example, if we are gating from audio stream to visual stream, then visual stream is Query and audio stream is Key and Value. The output has the same spatial-temporal dimension as Query. 

\begin{figure}
\begin{center}
\includegraphics[width=0.8\linewidth]{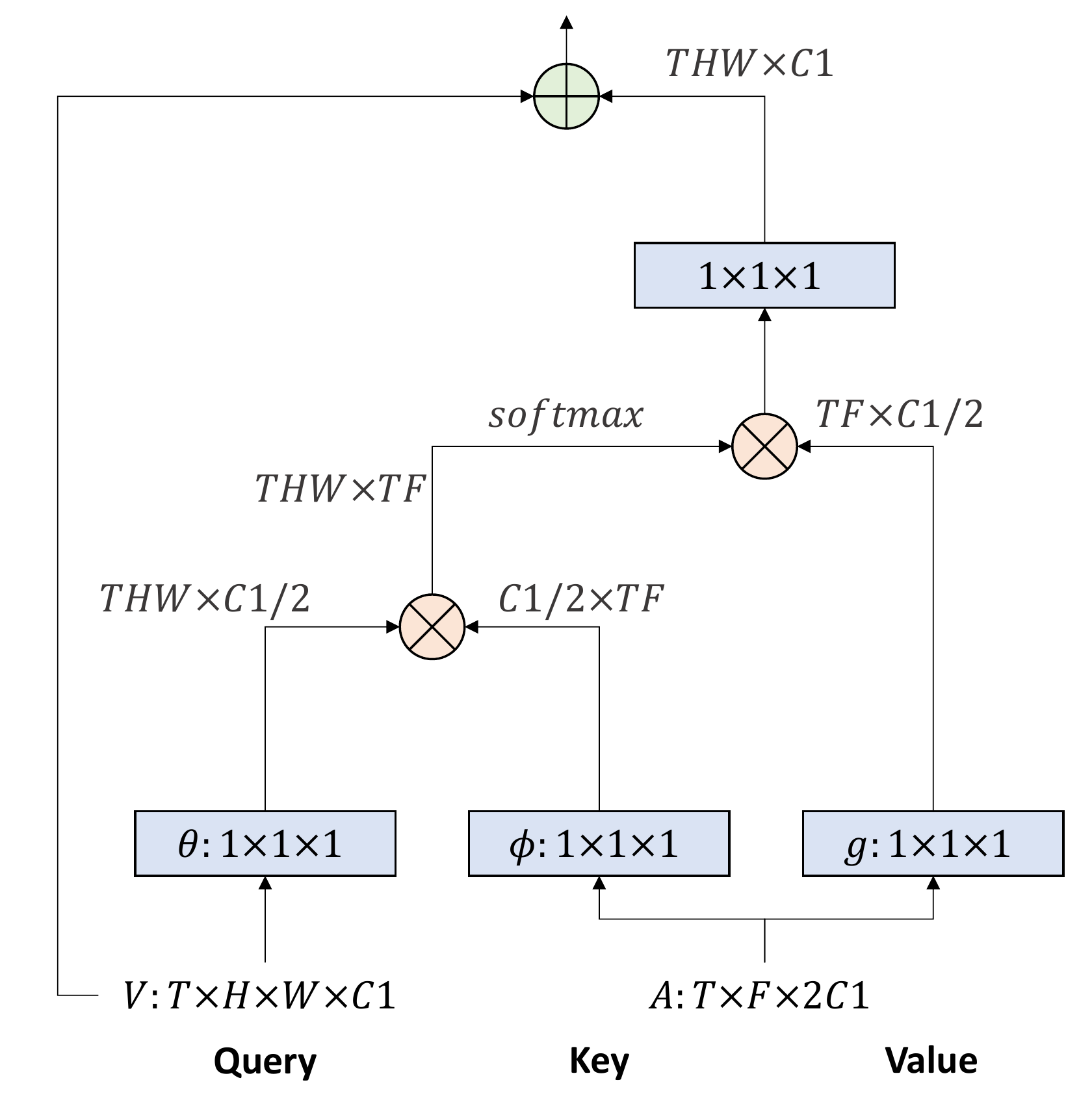}
\caption{{\bf NL-Gate Implementation}. Figure of the implementation of NL-Gate on visual stream. Visual features are the Query. The 2D Mid-Concatenation of visual and audio features is the Key and Value.}
\label{fig:NL-Gate}
\end{center}
\vspace{-3mm}
\end{figure}

Specifically, we use Non-Local gate in \cite{XiaolongWang18} as the implementation for Query-Key-Value attention mechanism. Details of the design are illustrated in fig.~\ref{fig:NL-Gate}. Similar to SE-gate, NL-Gate can be added with multiple directions and at multiple positions. We found that it works the best when added after block 4, with a 2-D concat of audio and RGB features as Key-Value and visual features as Query to gate the visual stream. 

\section{Additional Ablation Results}

\subsection{A strong oracle baseline}

In section~3.3, we presented the results on Gradient-Blending as an effective regularizer to train multi-modal networks. Here, we consider an additional strong baseline for the Kinetics, audio-RGB case. 

Suppose we have an oracle to choose the best modality (from audio, RGB and naive A/V) for each class. For example, for ``whistling" video, the oracle chooses naive A/V model as it performs the best among the three on ``whistling" in validation set. With this oracle, Top-1 video accuracy is 74.1\%, or 0.6\% lower than the offline G-Blend result.

\subsection{Training Accuracy}
In section~3.2, we introduced the overfitting problem of joint training of multi-modal networks. Here we include both validation accuracy and train accuracy of the multi-modal problems (Table~\ref{tab:of_naive_t_v}). We demonstrate that in all cases, the multi-modal networks are performing worse than their single best counterparts, while almost all of their train accuracy are higher (with the sole exception of OF+A, whose train accuracy is similar to audio network's train accuracy).

\begin{table}
	\centering
	\small
	\begin{tabular}{|c|c|c|c|}
	\hline
	Dataset & Modality & Validation V@1 & Train V@1 \\
	\hline
	\multirow{7}{4em}{Kinetics} & A & 19.7 & 85.9 \\
	 & RGB & 72.6 & 90.0 \\
	 & OF & 62.1 & 75.1 \\
	 & A + RGB & 71.4 & 95.6 \\
	 & RGB + OF & 71.3 & 91.9 \\
	 & A + OF & 58.3 & 83.2 \\
	 & A + RGB + OF & 70.0 & 96.5 \\
	\hline
	\multirow{3}{4em}{mini-Sport} & A & 22.1 & 56.1 \\
	& RGB & 62.7 & 77.6 \\
	& A + RGB & 60.2 & 84.2 \\
	\hline
	\end{tabular} 
	\caption{\textbf{Multi-modal networks have lower validation accuracy but higher train accuracy.} Table of Top-1 accuracy of single stream models and naive late fusion models. Single stream modalities include RGB, Optical Flow (OF), and Audio Signal (A). Its higher train accuracy and lower validation accuracy signal severe overfitting.} 
	\label{tab:of_naive_t_v}
	\vspace{-5mm}
\end{table}

\subsection{Early Stopping}

\begin{figure}[h]
\begin{center}
\includegraphics[width=0.7\linewidth]{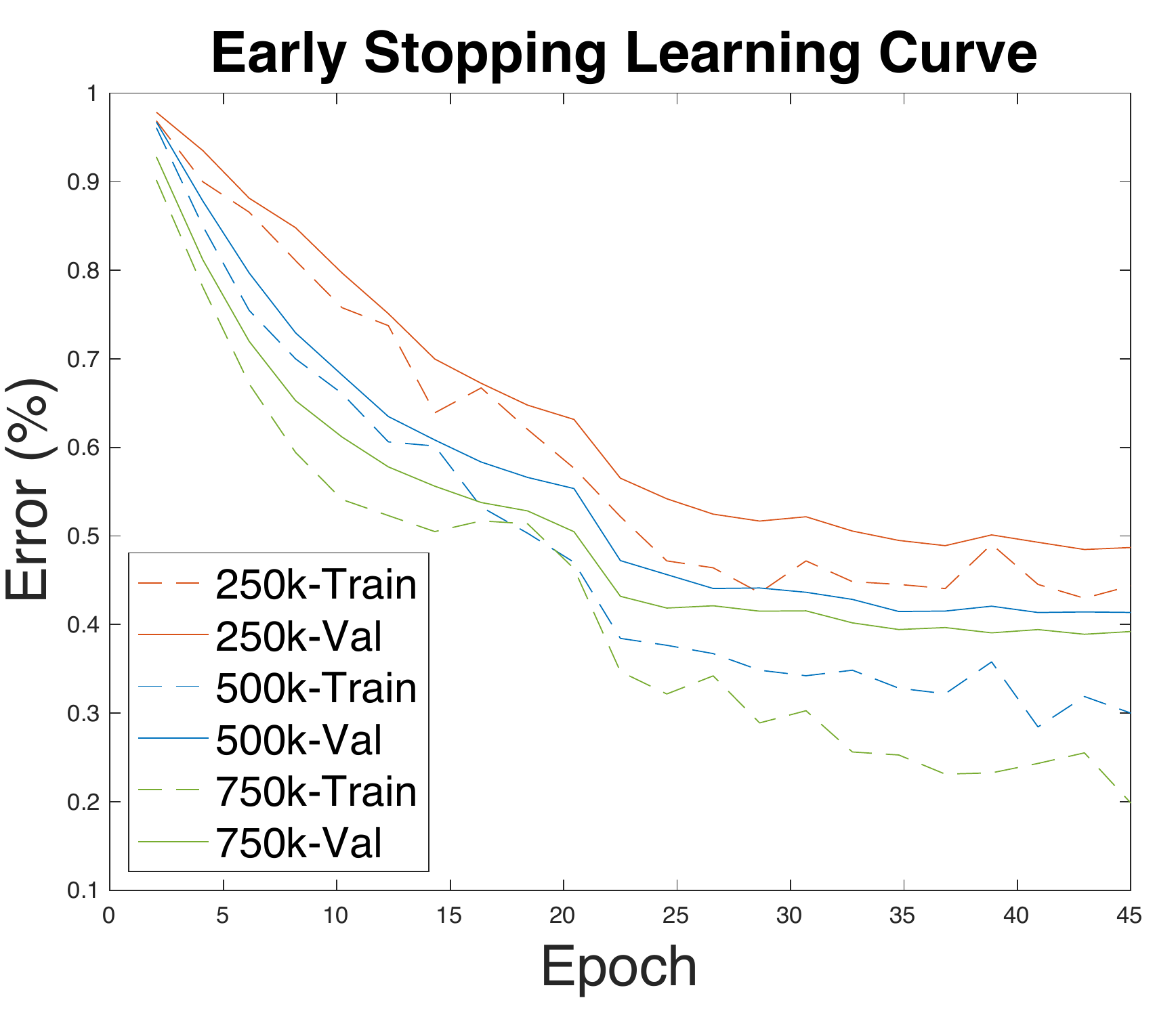}
\caption{{\bf Early stopping avoids overfitting but tends to under-fit.} Learning curves for three early stopping schedules we experiment. When we train the model with less number of iterations, the model does not overfit, but the undesirable performance indicates an under-fitting problem instead.}
\label{fig:early_stop_ablation}
\end{center}
\end{figure}

In early stopping, we experimented with three different stopping schedules: using 25\%, 50\% and 75\% of iterations per epoch. We found that although overfitting becomes less of a problem, the model tends to under-fit. In practice, we still found that the 75\% iterations scheduling works the best among the three, though it's performance is worse than full training schedule that suffers from overfitting. We summarize their learning curves in fig.~\ref{fig:early_stop_ablation}.

\subsection{Additional Qualitative Analysis}

In section 3.3 we presented the qualitative analysis of G-Blend's performance compared with RGB model performance (fig.6). We expand the analysis and provide more details in this section.

\begin{figure*}[h]
 \centering
 \captionsetup{font=footnotesize}
 \begin{subfigure}[b]{0.35\linewidth}
    {\includegraphics[width=\linewidth]{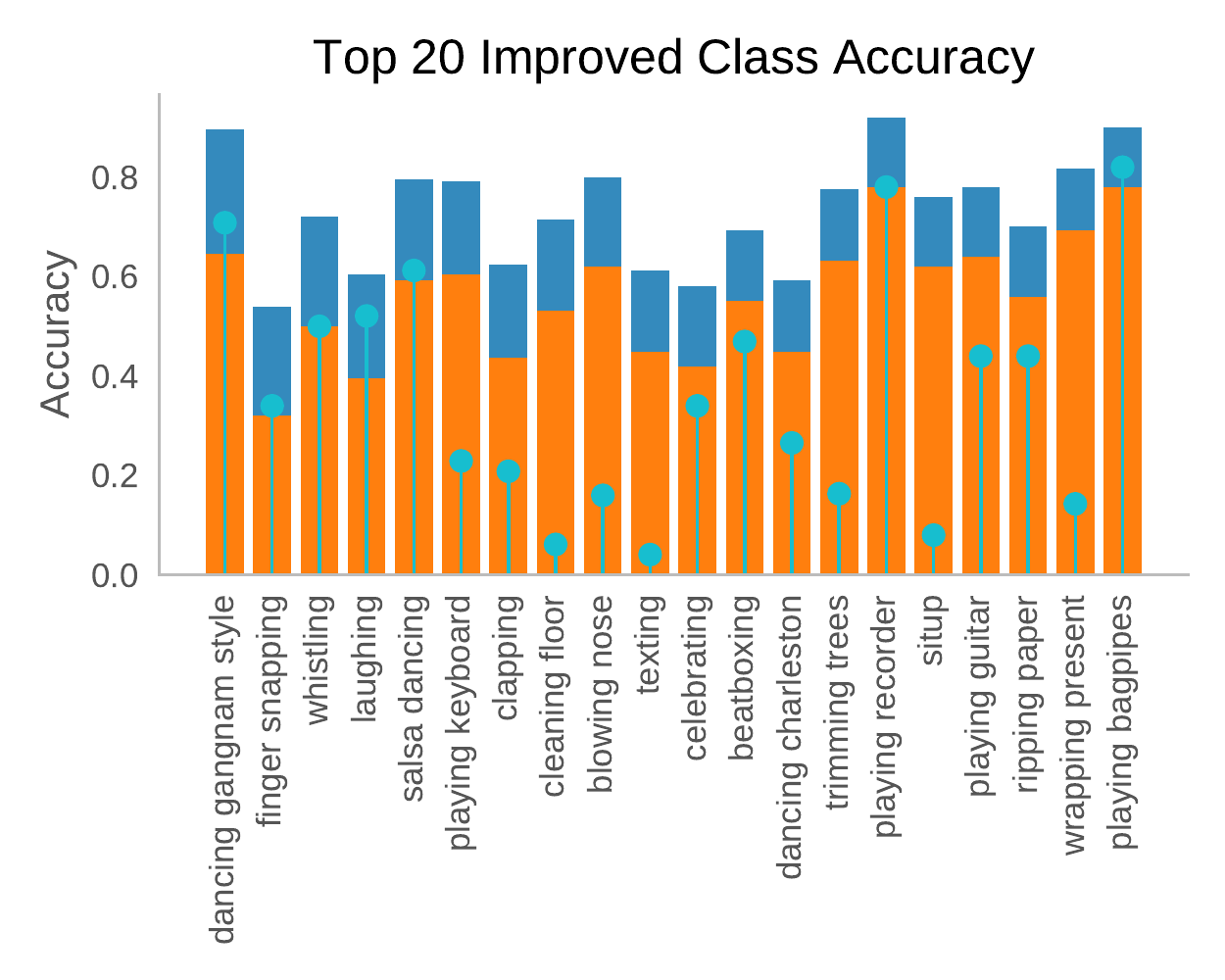}}
 \end{subfigure}
 \begin{subfigure}[b]{0.35\linewidth}
    {\includegraphics[width=\linewidth]{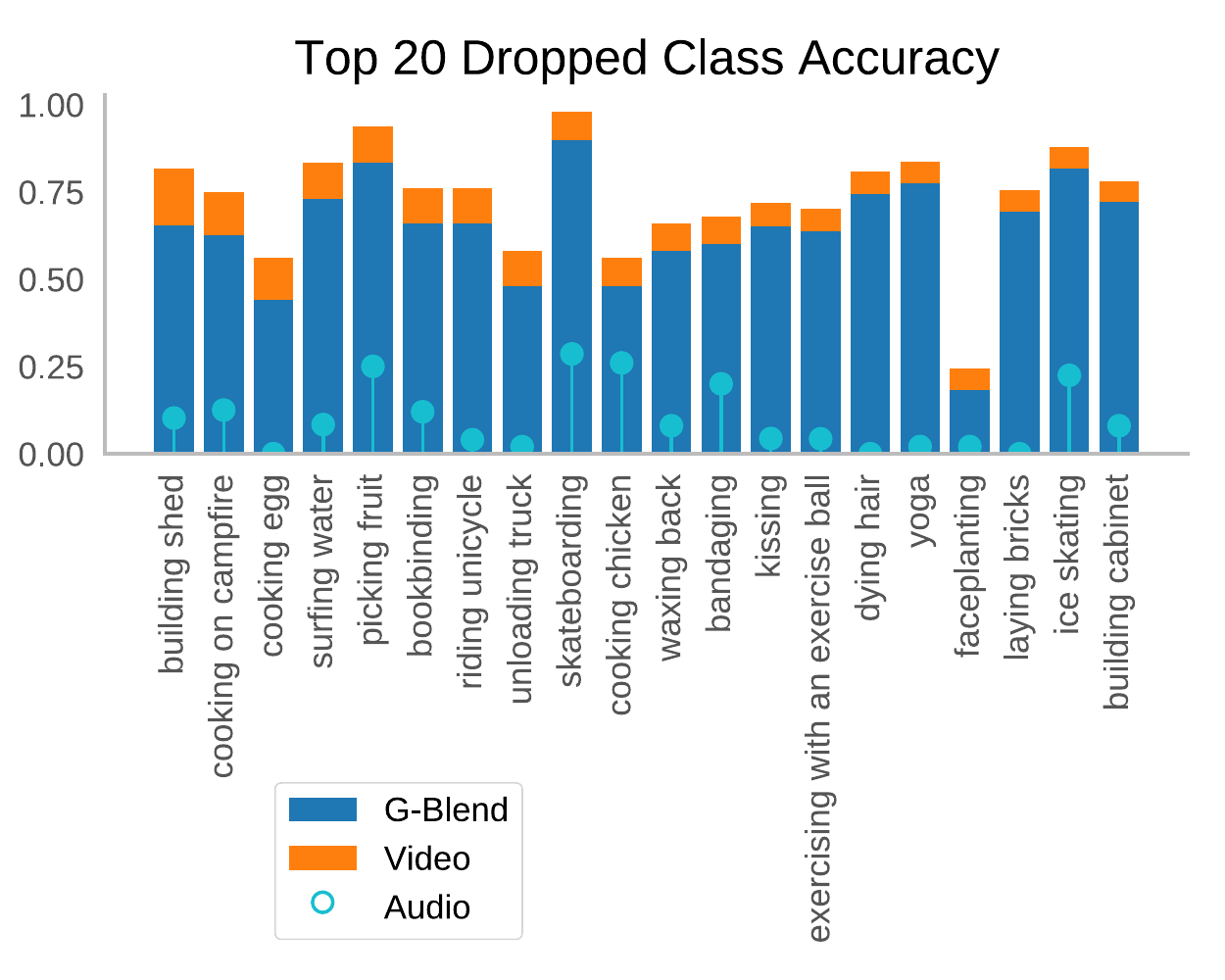}}
 \end{subfigure}
 \vspace{-3mm} 
 \caption{\textbf{Top-Bottom 20 classes based on improvement of G-Blend to RGB model.} The improved classes are indeed audio-relevant, while those have performance drop are not very audio semantically-related.}
 \vspace{-6mm} 
 \label{fig:gb_top_bot_20}
\end{figure*}

We first expand the analysis to compare the top-20 and bottom-20 improved classes of G-Blend versus RGB model (fig.~\ref{fig:gb_top_bot_20}). This is a direct extension of fig.6. It further confirms that classes that dropped are indeed not very semantically relevant in audio, and in many of those classes, the audio model's performance is almost 0. 

\begin{figure*}[h]
 \centering
 \captionsetup{font=footnotesize}
 \begin{subfigure}[b]{0.35\linewidth}
    {\includegraphics[width=\linewidth]{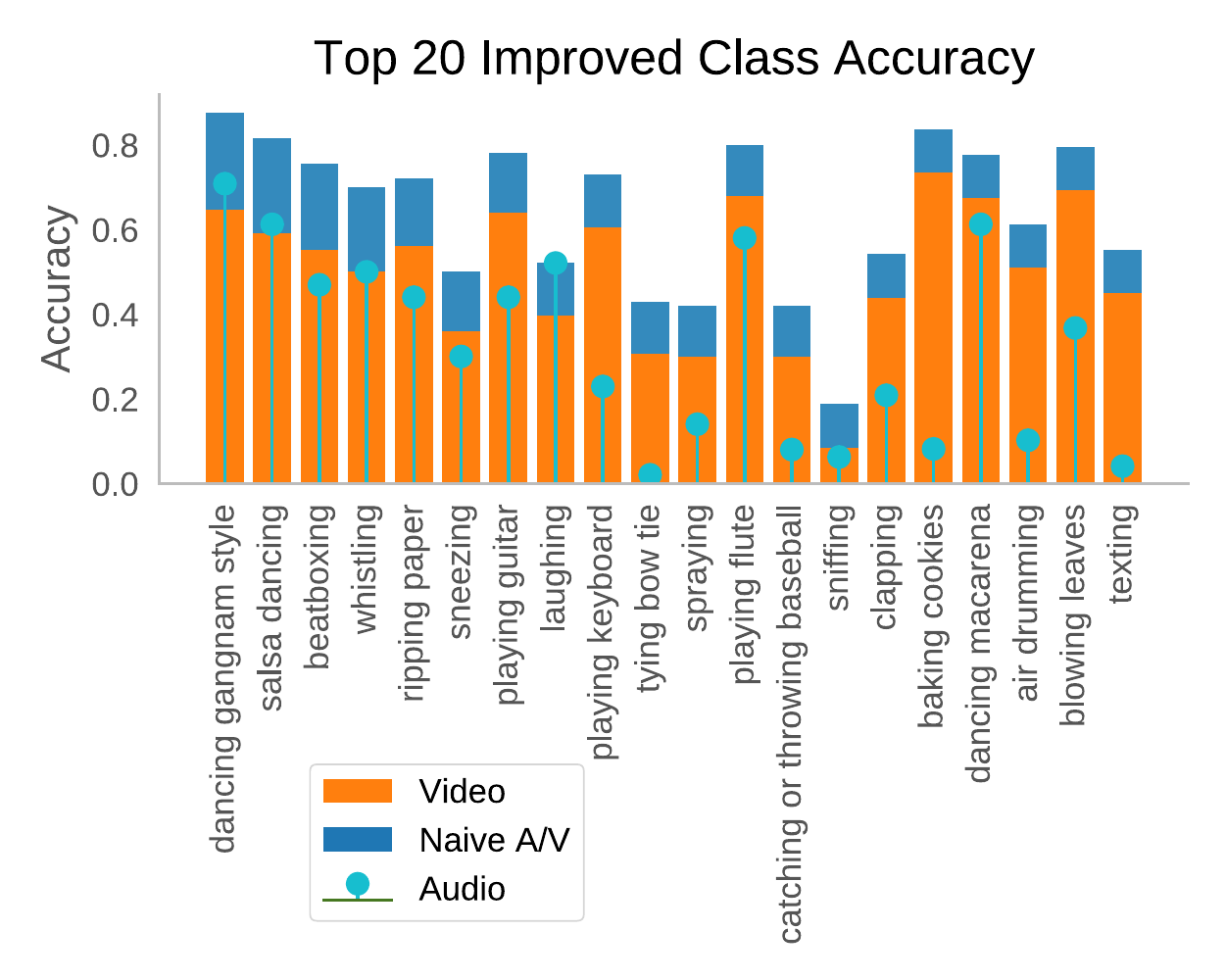}}
 \end{subfigure}
 \begin{subfigure}[b]{0.35\linewidth}
    {\includegraphics[width=\linewidth]{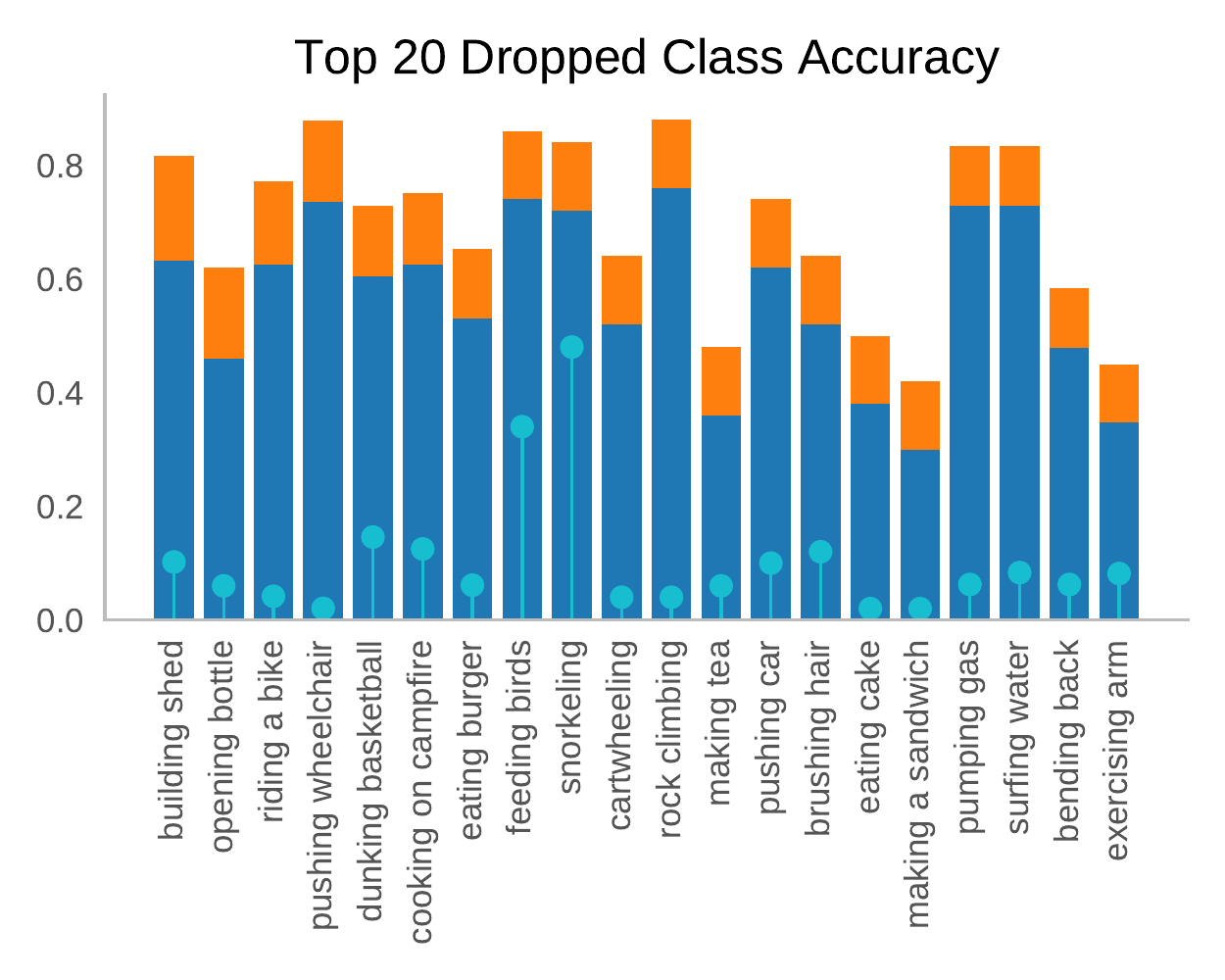}}
 \end{subfigure}
 \vspace{-3mm} 
 \caption{\textbf{Top-Bottom 20 classes based on improvement of naively trained audio-visual to RGB model.} The improvement tends to be smaller than that of G-B counterpart and the drop is more significant. More interesting, in some classes, the naively trained A/V model performs worse than audio signal.}
 \vspace{-6mm} 
 \label{fig:naive_top_bot_20}
\end{figure*}

We further extends the analysis to compare naively trained audio-visual model with RGB-only model (fig.~\ref{fig:naive_top_bot_20}). We note that the improvement for top-20 classes is smaller than that of G-B and for bot-20 classes the drop is mroe significant. Moreover, we note that in some bot-20 classes like snorkeling or feeding bird, where the sound of breathing and birds is indeed relevant, naively trained A/V model is not performing well. For these classes, audio model achieves decent performance. We further note that interestingly, for laughing, although naive A/V model outperforms RGB model, it is worse than audio-only model. And only with G-Blend, it benefits from both visual and audio signals, performing better than both.   

Finally, we compare the top-20 and bot-20 classes where G-Blend has the most improvement/ drop with naively trained A/V model. We note that the gains in improved classes are much larger than the decrease in dropped classes. 

\begin{figure*}[h]
 \centering
 \captionsetup{font=footnotesize}
 \begin{subfigure}[b]{0.35\linewidth}
    {\includegraphics[width=\linewidth]{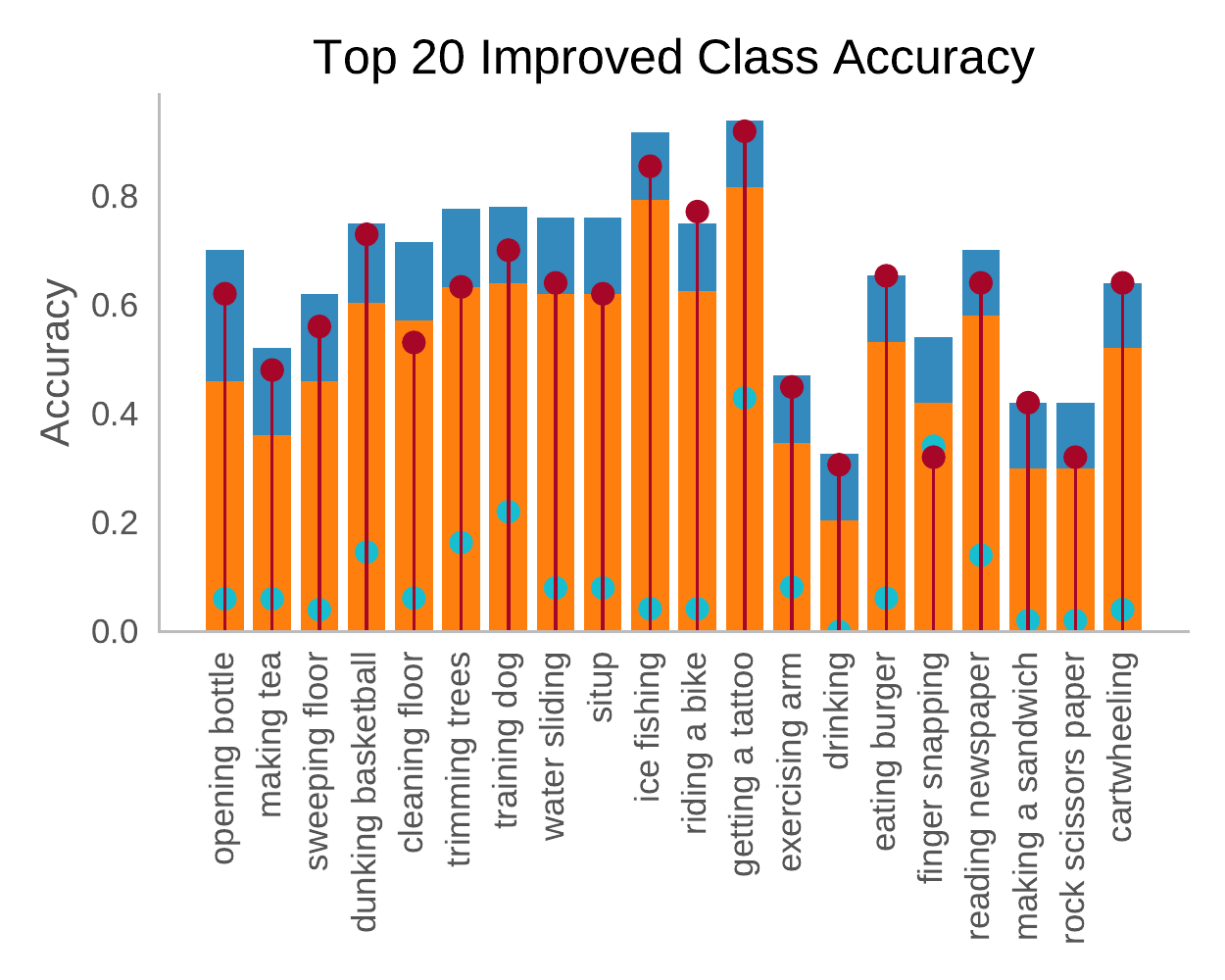}}
 \end{subfigure}
 \begin{subfigure}[b]{0.35\linewidth}
    {\includegraphics[width=\linewidth]{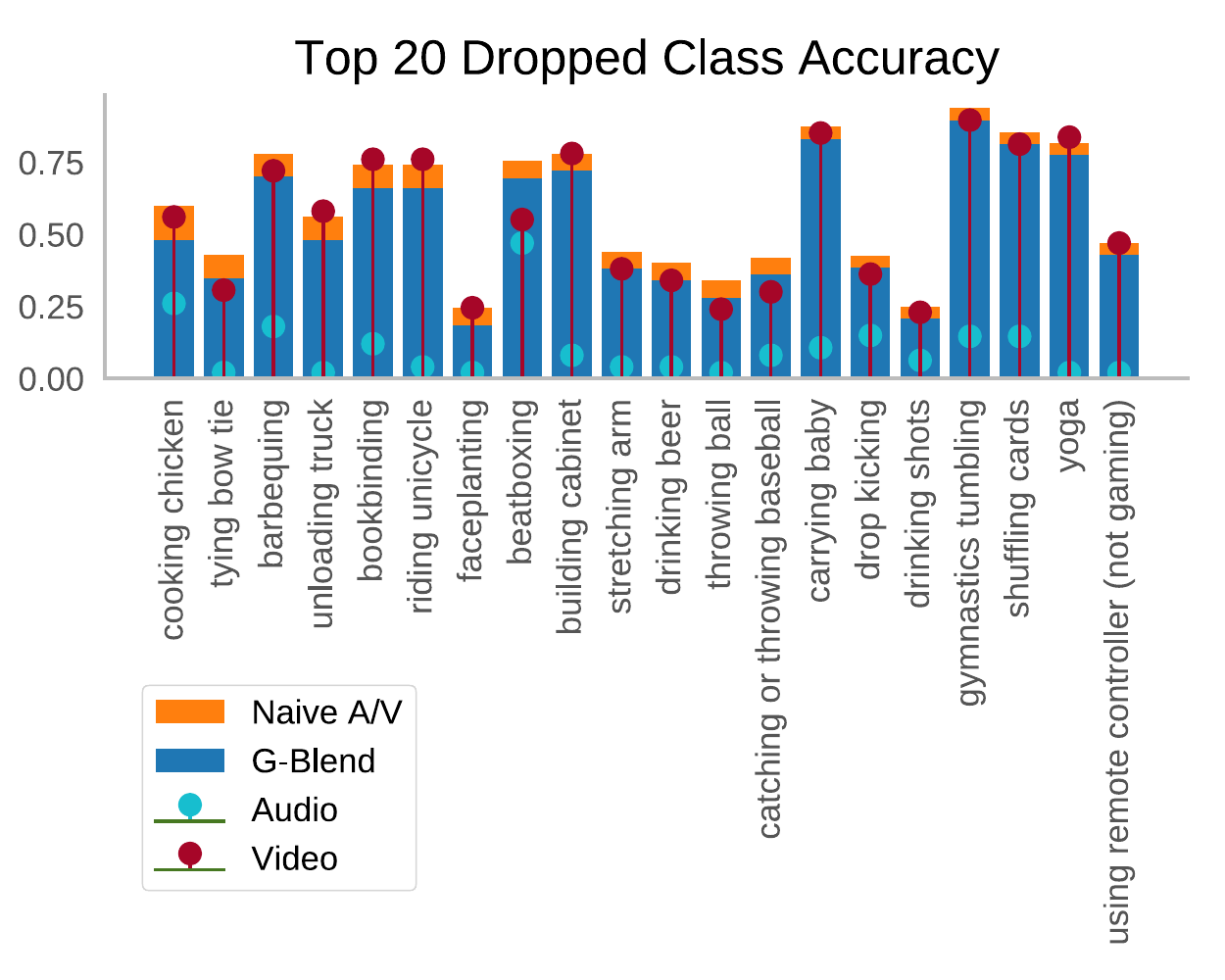}}
 \end{subfigure}
 \vspace{-3mm} 
 \caption{\textbf{Top-Bottom 20 classes based on improvement of G-Blend to Naive audio-visual model.} We note that the gain is much more significant than drop.}
 \vspace{-6mm} 
 \label{fig:gbnaive_top_bot_20}
\end{figure*}

\end{document}